\documentclass[10pt,twocolumn,letterpaper]{article}

\usepackage[pagenumbers]{cvpr}

\usepackage{tabularx}
\usepackage{pifont}
\newcommand{\cmark}{\ding{51}}
\newcommand{\xmark}{\ding{55}}

\definecolor{cvprblue}{rgb}{0.21,0.49,0.74}
\usepackage[pagebackref,breaklinks,colorlinks,allcolors=cvprblue]{hyperref}

\title{SceneBench: A Hierarchical Benchmark for Vision-Language Understanding of 3D Scenes}

\author{
Anubhav Khanal$^{1}$ \qquad
Prabigya Acharya$^{1}$ \qquad
Roshni Poudel$^{1}$ \qquad
Sujan Kapali$^{1}$ \qquad
Bigyan Bhatta$^{1}$ \\
Pramish Paudel$^{2}$ \qquad
Francois Rameau$^{1,3}$ \qquad
Danda Pani Paudel$^{1,2}$ \\[0.5em]
$^{1}$Nepal Applied Mathematics and Informatics Institute for research (NAAMII), Nepal \\
$^{2}$INSAIT, Sofia University, Bulgaria \\
$^{3}$State University of New York at Stony Brook, Korea
}

\begin{document}
\maketitle
\begin{figure*}[t]
    \centering
    \includegraphics[width=\textwidth]{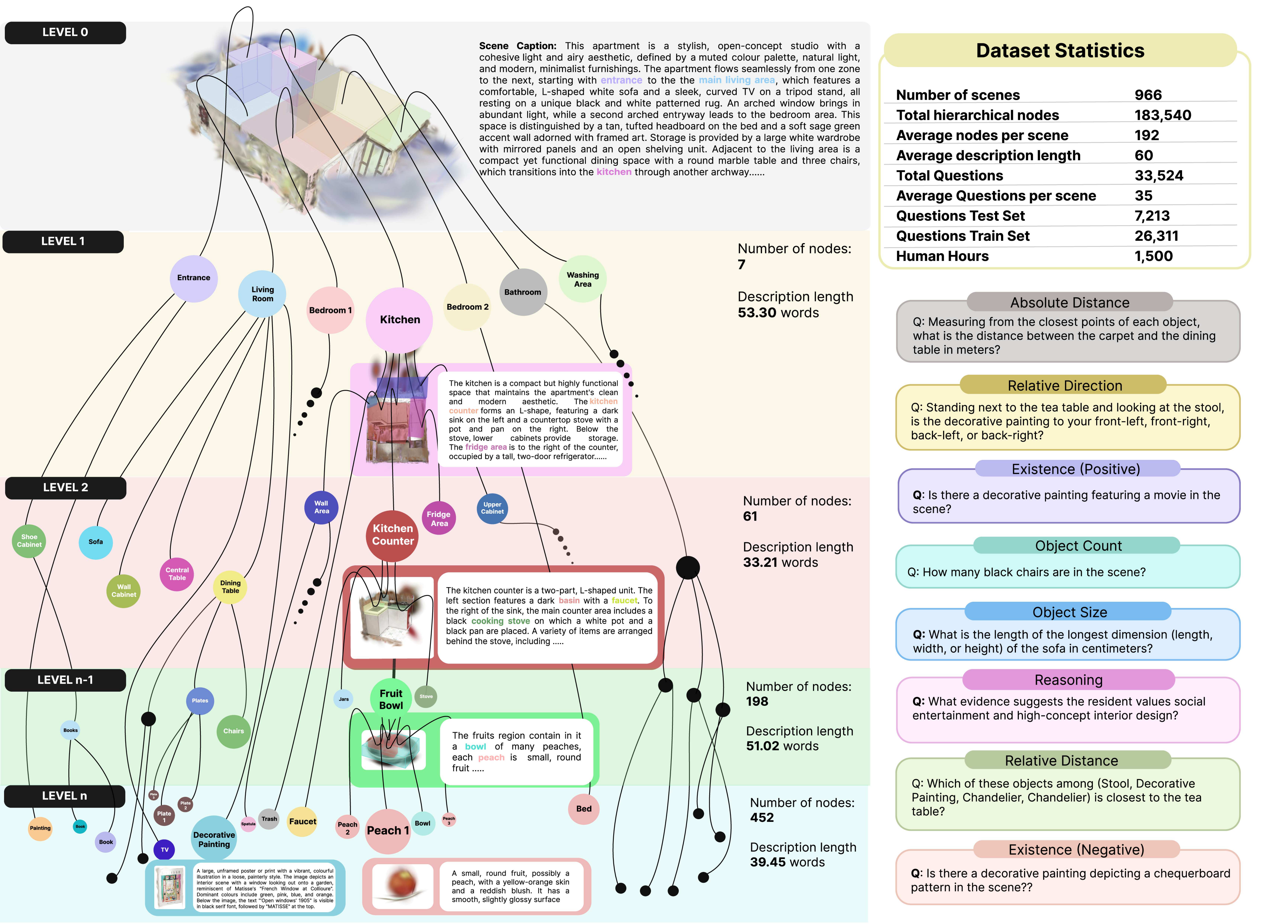}
    \caption{
\textbf{Overview of dataset construction and statistics.}
\textbf{(Left)} The hierarchical scene description structure:
each level $(n-1)$ description grounds all objects within its
spatial scope by referencing finer-grained level $(n)$
descriptions, all the way to the leaf nodes, where each node
corresponds to an individual object in the 3D Gaussian
Splatting scene.
\textbf{(Top Right)} Dataset statistics summarizing the scale
of the corpus across scenes, hierarchical nodes, description
length, and question distribution.
\textbf{(Bottom Right)} The seven question types spanning
existence (with positive and negative variants), five
spatially grounded tasks (absolute distance, relative
distance, relative direction, object count, and object size),
and one multi-step reasoning task, collectively designed
to evaluate fine-grained spatial and semantic understanding
of 3D scenes.
}
    \label{fig:Teaser}
\end{figure*}

\begin{abstract}
Vision-language models have achieved strong performance on 2D image understanding, but their ability to reason about 3D environments remains limited. Progress in spatial intelligence is hindered by limitations in existing benchmarks. First, many 3D datasets rely on point clouds, which capture geometry but discard rich visual appearance such as texture, text, and materials. Second, annotations typically treat objects in isolation and overlook the hierarchical organization of real environments into scenes, rooms, functional areas, and object groups. Third, current evaluation tasks focus largely on recognition or localization and do not test multi-step spatial reasoning within complex scenes.

In this context, we introduce \textbf{SceneBench}, a benchmark of \textbf{966 photorealistic 3D scenes} reconstructed with Gaussian Splatting and densely annotated with hierarchical semantics spanning scenes, rooms, functional areas, object groups, and individual objects. These annotations are produced through a human-in-the-loop pipeline combining vision-language models with roughly \textbf{1,500 human-hours} of iterative refinement and verification, producing over \textbf{183K} annotated nodes with textual descriptions and 3D bounding boxes. Building on this representation, we define three evaluation tasks: \emph{Existence-Based Questions} probing object attributes, \emph{Spatial Intelligence Questions} covering counting, size comparison, distance, and directional relations, and \emph{Grounded Question-Reasoning-Answer (QRA)} triplets requiring multi-step reasoning across semantic levels. Experiments with state-of-the-art vision-language models show that while models perform well on basic recognition tasks (e.g., up to \textbf{85\%} accuracy for detection), performance drops substantially on hierarchical and compositional reasoning (e.g., down to \textbf{60\%} for counting), revealing limitations not captured by existing benchmarks. SceneBench provides a realistic testbed for developing and evaluating models capable of fine-grained spatial reasoning in photorealistic 3D environments. Our dataset and benchmark will be made publicly available upon acceptance.
\end{abstract}
    
\section{Introduction}
Recent multimodal vision-language models can associate language with visual observations, but their performance degrades when moving from image-level analysis to 3D scenes~\cite{chen2024spatialvlm}. Consequently, they remain unsuitable for real-world deployment, where intelligent agents require spatial intelligence: the ability to jointly reason about geometry, appearance, scale, and object relationships in 3D spaces. Evaluating and improving these capabilities therefore requires benchmarks that capture both the visual detail and structural organization of real-world scenes.

However, existing benchmarks do not meet these requirements. Most rely on 2D images or video, which lack a persistent 3D structure and do not support reasoning across viewpoints. Others rely on geometry-only 3D representations such as point clouds \cite{azuma2022scanqa, chen2020scanrefer, chen2021scan2cap, ma2022sqa3d, Chen_2024_CVPR, huang20253d}, which capture coarse spatial layout but discard important visual cues such as text, texture, material properties, and lighting. In contrast, high-fidelity 3D representations such as Gaussian Splatting (GS)~\cite{kerbl20233d} preserve detailed appearance across arbitrary viewpoints while maintaining explicit 3D structure, making them well suited for holistic spatial reasoning in realistic environments. However, few GS-based datasets provide rich semantic annotations, largely due to the lack of scalable annotation pipelines.

Beyond representation, current benchmarks also suffer from two additional limitations. First, most annotations are limited to individual object categories, ignoring the hierarchical organization of real environments into scenes, rooms, functional areas, object groups, and parts \cite{azuma2022scanqa}. By treating objects in isolation, these datasets fail to capture the semantic structure of real-world environments. Second, existing benchmarks primarily evaluate recognition or localization queries, which do not test the multi-step reasoning required to navigate a scene, resolve ambiguities, and identify fine-grained targets based on both visual and spatial cues.

To address these gaps, we introduce \textbf{SceneBench}, a new benchmark for immersive and hierarchical 3D vision-language understanding, illustrated in \cref{fig:Teaser}. SceneBench consists of \textbf{966 photorealistic 3D scenes} with \textbf{183K} text-annotated nodes arranged hierarchically, represented using 3D Gaussian Splatting and densely annotated with semantics spanning scenes, rooms, functional areas, object groups, and individual objects. Our contributions are twofold:
\begin{itemize}
    \item \textbf{A hierarchical, tree-based vision-language representation} that organizes semantics from scene-level concepts down to object parts and attributes. Each leaf node corresponds to an object, while inner nodes represent hierarchical regions grounded in 3D with both textual descriptions and spatial localization.
    
    \item \textbf{A benchmark for spatial intelligence in photorealistic 3D scenes} that evaluates both core spatial reasoning (e.g., relative arrangements, counts, distances) and advanced multi-step reasoning requiring cross-object references and fine-detail analysis. Furthermore, we show that existing methods struggle on our benchmark, and provide a detailed analysis.
\end{itemize}

\section{Related Work}
\label{sec:related_work}

The field of 3D scene understanding has transitioned from task-specific architectures operating on point clouds~\cite{qi2017pointnet++, dai2017scannet} toward unified multimodal representations. This shift occurred with the emergence of neural scene representations such as NeRF~\cite{mildenhall2021nerf} and 3D GS~\cite{kerbl20233d}. Building on this idea, several works have distilled language and semantic information into these representations to support open-vocabulary queries and zero-shot reasoning in 3D environments~\cite{qin2024langsplat, zhou2024feature, thai2025splattalk, ye2024gaussian, cen2025segment, kerr2023lerf, cheng2025occamslgsefficientapproach, wu2024opengaussian, lu2025segment, liao2025spc, wang2025plgs}.
Recent datasets have followed the same direction by offering large-scale 3DGS scenes with dense object-level annotations~\cite{spatialverse2025interiorgs}. However, such datasets remain largely limited to flat object structures, discarding the multi-level semantic hierarchies and fine-grained visual details necessary for complex spatial intelligence.

Hierarchical 3D Scene Graphs structure environments into multiple levels of abstraction. 
Early works introduced hierarchical scene graphs to support long-range navigation and overcome LLM context limits~\cite{armeni20193d, werby2024hovsg, werby2025keysg}.
More recently, datasets have leveraged scene graphs to automatically generate large multimodal perception suites and dense grounded annotations~\cite{Yang_2025_CVPR, wang2024embodiedscan, jia2024sceneverse}.

Some approaches go beyond object-level grounding and incorporate fine-grained parts~\cite{yeshwanth2025excap3d}.
A quantitative comparison of these 3D visual grounding datasets is provided in \cref{tab:related_datasets}.

While algorithmic scene graph generation over 3D Gaussians is actively being explored~\cite{wang2025gaussiangraph3dgaussianbasedscene}, explicit, multi-level relational hierarchies natively embedded within a dataset benchmark remain limited to mesh or point-cloud environments.

To leverage structured 3D environments, recent research has explored grounding large language models (LLMs) for spatial reasoning and question answering, as summarized in \cref{tab:dataset_comparison}. Early benchmarks introduced object-grounded queries over point clouds and egocentric views~\cite{azuma2022scanqa, ma2023sqa3d}. Subsequent work expanded these tasks toward multi-view and compositional reasoning~\cite{hong20233d, zhang2025open3dvqabenchmarkcomprehensivespatial}. Other studies emphasize the importance of explicit cognitive mapping for spatial reasoning, proposing benchmarks based on 2D videos or large image collections~\cite{yang2025thinking, deng2025internspatialcomprehensivedatasetspatial, liu2025miragemultimodalbenchmarkspatial}. In parallel, advanced reasoning frameworks incorporate reinforcement learning and internal visualization-of-thought to improve multi-step embodied task performance~\cite{huang20253dr1, huang2025surprised}. Despite these advances, no existing dataset combines photorealistic 3D rendering with densely grounded and structured question answering, leaving a gap between visual fidelity and spatial reasoning.

To address these limitations, we introduce SceneBench, the first dataset combining the high-fidelity visual rendering of 3DGS with densely annotated hierarchical scene graphs. By explicitly modeling multi-scale relational dependencies and defining Existence, Spatial Intelligence, and Grounded Question\allowbreak-Reasoning\allowbreak-Answer tasks, SceneBench establishes a benchmark that connects simple 3D localization with higher-level spatial reasoning.

\begin{table}[t]
\centering
\caption{Comparison of 3D scene understanding datasets. \textbf{Q.} indicates whether the dataset includes a question-answering component. \textbf{Descr.} denotes the semantic granularity of the environment annotations (S=Scene, R=Room, G=Groups, O=Object, P=Part), and \textbf{Gr.} indicates whether queries are explicitly linked to 3D geometry. \textbf{Q. Types} abbreviations: Sp=Spatial, Ct=Counting, AO=Appearance Order, RP=Route Planning, Ms=Measurement, Dr=Directional, At=Attributes, Ex=Existence, Tm=Temporal, Cb=Combination, Rl=Relation, CSR=Common Sense Reasoning; MS=Multi-step.}
\label{tab:dataset_comparison}
\small
\setlength{\tabcolsep}{2pt}
\renewcommand{\arraystretch}{1.05}
\resizebox{\linewidth}{!}{%
\begin{tabular}{l c l l c l l}
\toprule
\textbf{Dataset} & \textbf{Q.} & \textbf{3D Repr.} & \textbf{Descr.} & \textbf{Gr.} & \textbf{Q. Types} & \textbf{Reason.}\\
\midrule
\itshape
Thinking in Space \cite{yang2025thinking} & \cmark & Vid & S & \xmark & Sp, Ct, AO, RP, Ms & \cmark (Sp) \\
ExCap3D \cite{yeshwanth2025excap3d} & \xmark & PC & O, P & \cmark & N/A & \xmark \\
3D-R1 \cite{huang20253dr1} & \cmark & Img, PC, Txt & S & \cmark & Sp, Dr, At, Ex, Ct & \cmark (CoT) \\
3D-GRAND \cite{yang20253dgrandmillionscaledataset3dllms} & \xmark & PC & S & \cmark & Sp, Dr, Ex, At, Ct & \cmark \\
3DMV-VQA \cite{hong20233d} & \cmark & MV-Img & S, R & \cmark & Ex, Ct, Sp, Ms & \cmark \\
ScanQA \cite{azuma2022scanqa} & \cmark & PC & S, O & \cmark & At, Sp, Dr, Ex, Ct & \xmark \\
InternSpatial \cite{deng2025internspatialcomprehensivedatasetspatial} & \cmark & Img & S, O & \cmark & Sp, Ct, Tm, Ex & \xmark \\
MIRAGE \cite{liu2025miragemultimodalbenchmarkspatial} & \cmark & Img & O & \xmark & Cb, Ct, Rl & \xmark \\
SceneVerse \cite{jia2024sceneverse} & \xmark & PC & S, O & \cmark & N/A & \xmark \\
SURPRISE3D \cite{huang2025surprised} & \cmark & PC & O & \cmark & Sp, CSR & \cmark \\
\upshape
\textbf{Ours} & \textbf{\cmark} & \textbf{3DGS} & \textbf{S, R, G, O} & \textbf{\cmark} & \textbf{Sp, Ex, Ct, Dr} & \textbf{\cmark (MS)} \\
\bottomrule
\end{tabular}
}
\end{table}

\begin{table}[t]
\centering
\caption{Comparison of 3D visual grounding datasets that pair 3D geometry with natural language descriptions. \textbf{Objects}: number of unique 3D objects; \textbf{Descriptions}: total number of language descriptions; \textbf{Avg. Length}: average word count per description; \textbf{Scene Type}: 3D representation format.}
\label{tab:related_datasets}
\footnotesize
\setlength{\tabcolsep}{3pt}
\begin{tabular}{@{}l r r c c@{}}
\toprule
\textbf{Dataset} & \textbf{Objects} & \textbf{Descr.} & \textbf{Avg.\ Len.} & \textbf{Scene Type}\\
\midrule
ScanRefer~\cite{chen2020scanrefer} & 11,046 & 51,583 & 20.27 & 3D Scan \\
ReferIt3D~\cite{achlioptas2020referit_3d} & 96,654 & 130,364 & 3.51 & Image \\
RioRefer~\cite{miyanishi2024cross3dvgcrossdataset3dvisual} & 31,801 & 63,602 & 14.78 & Image \\
ArkitSceneRefer~\cite{kato-etal-2023-arkitscenerefer} & 15,553 & 15,553 & 14.43 & 3D Scan \\
SceneVerse~\cite{jia2024sceneverse} & 1.5M & 2.5M & - & Point Cloud \\
3D-Grand~\cite{yang20253dgrandmillionscaledataset3dllms} & 6.2M & 6.2M & - & Point Cloud \\
\textbf{Ours} & \textbf{128,966} & \textbf{183,540} & \textbf{60} & \textbf{3DGS} \\
\bottomrule
\end{tabular}
\end{table}
    
\begin{figure*}[t]
    \centering
    \includegraphics[width=\textwidth]{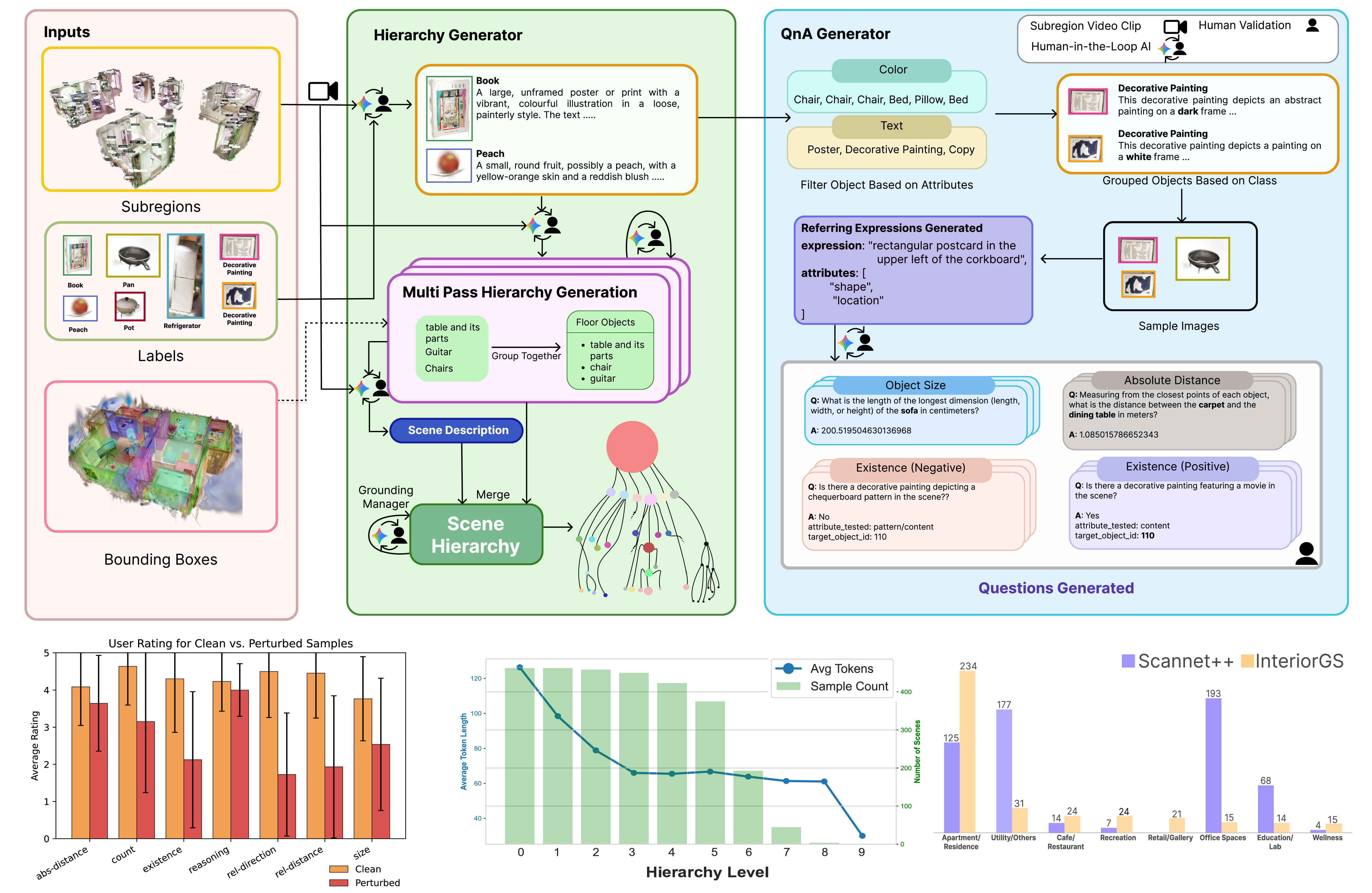}
    \caption{\textbf{Hierarchy and Question Generation Process} The top diagram illustrates our pipeline for generating scene hierarchies and QA pairs from 3DGS scenes. The bottom row presents dataset statistics: (\textbf{left}) user ratings comparing clean versus perturbed samples across question types; (\textbf{center}) average token length (blue) and scene count (green) at each hierarchy depth level; (\textbf{right}) distribution of scenes across room categories for ScanNet++ and InteriorGS subsets.}
    \label{fig:flow}
\end{figure*}

\section{Dataset}
To support more granular 3D scene understanding, we propose a tree-based representation for vision-language annotations, as illustrated in \cref{fig:Teaser}.
Each leaf node in the tree corresponds to an object, and inner nodes correspond to a hierarchy of objects, paired with a 3D-grounded textual description. When a node appears within the description of another, it is grounded as a constituent part of that inner node.

The dataset comprises 966 scenes containing 584 real-world scenes from ScanNet++~\cite{yeshwanth2023scannethighfidelitydataset3d} and 382 synthetic scenes from InteriorGS~\cite{spatialverse2025interiorgs}. All scenes are represented as 3D Gaussian Splats. For ScanNet++, we utilize the splats provided by SceneSplat~\cite{li2025scenesplatgaussiansplattingbasedscene}; for InteriorGS, the scenes are natively available in the Gaussian Splatting format. 

Through extensive manual curation and multi-stage verification, in which all model calls are checked for errors and iteratively refined, we developed a human-in-the-loop pipeline (approximately 1,500 human-hours in total) to generate highly detailed, accurate free-form text descriptions and a hierarchical scene structure. This pipeline consists of two primary components: Object Description Generation and Object Hierarchy Generation, as illustrated in \cref{fig:flow}.

\paragraph{\textbf{Object Description Generation}}
To generate descriptions of objects in the scene, we employ a Vision-Language Model (VLM), Gemini 3.0, providing it with the semantic labels of objects, the count per label (ensuring no unique instance is missed), and a video of the scene with overlaid instance IDs to explicitly ground the visual data. The source of the visual input depends on the scene type. For ScanNet++~\cite{yeshwanth2023scannethighfidelitydataset3d}, we use the ground-truth videos captured during the scanning process. For InteriorGS~\cite{spatialverse2025interiorgs}, we automatically generate camera trajectories and render images along these paths to create synthetic videos.

To handle the high complexity of InteriorGS scenes, we partition large scenes into subregions using dataset room metadata; if a room remains too complex, we manually divide it into intuitive sub-sections. During inspection of initial outputs, we observed that high-density clusters (e.g., a stack of books) often suffered from spatial inaccuracies, such as misattributing one object's description to another. To resolve these mapping errors and provide needed visual context, we supplied multi-view images for those dense regions. Finally, we checked for missing or incorrectly classified descriptions and re-ran the generation pipeline for those objects to ensure complete and accurate coverage.

\paragraph{\textbf{Object Hierarchy Generation}}
To model the global context, we organize individual objects into a multi-level hierarchy where the topmost node represents the entire scene and the leaf nodes represent individual objects. Following prior work~\cite{jia2024sceneverse}, we first establish structural relationships, including attached to (e.g., ceiling lights), hanging from (e.g., wall art), supported by, contained by, and grouped with (e.g., a single book within a collection).
This initial structural hierarchy is then verified by a human and refined by a VLM. By providing the model with the structural tree and the scene video, we prompt it to identify and add functional areas (e.g., "dining area" or "workspace"), transitioning the hierarchy from purely spatial to semantic. This refinement is performed in multiple passes to ensure logical consistency. Finally, to ensure that every description is grounded, we perform a final pass in which the model revises parent node descriptions to explicitly reference all underlying child nodes.

\section{Benchmarking Tasks}
\begin{figure*}[t]
    \centering
    \includegraphics[width=\textwidth]{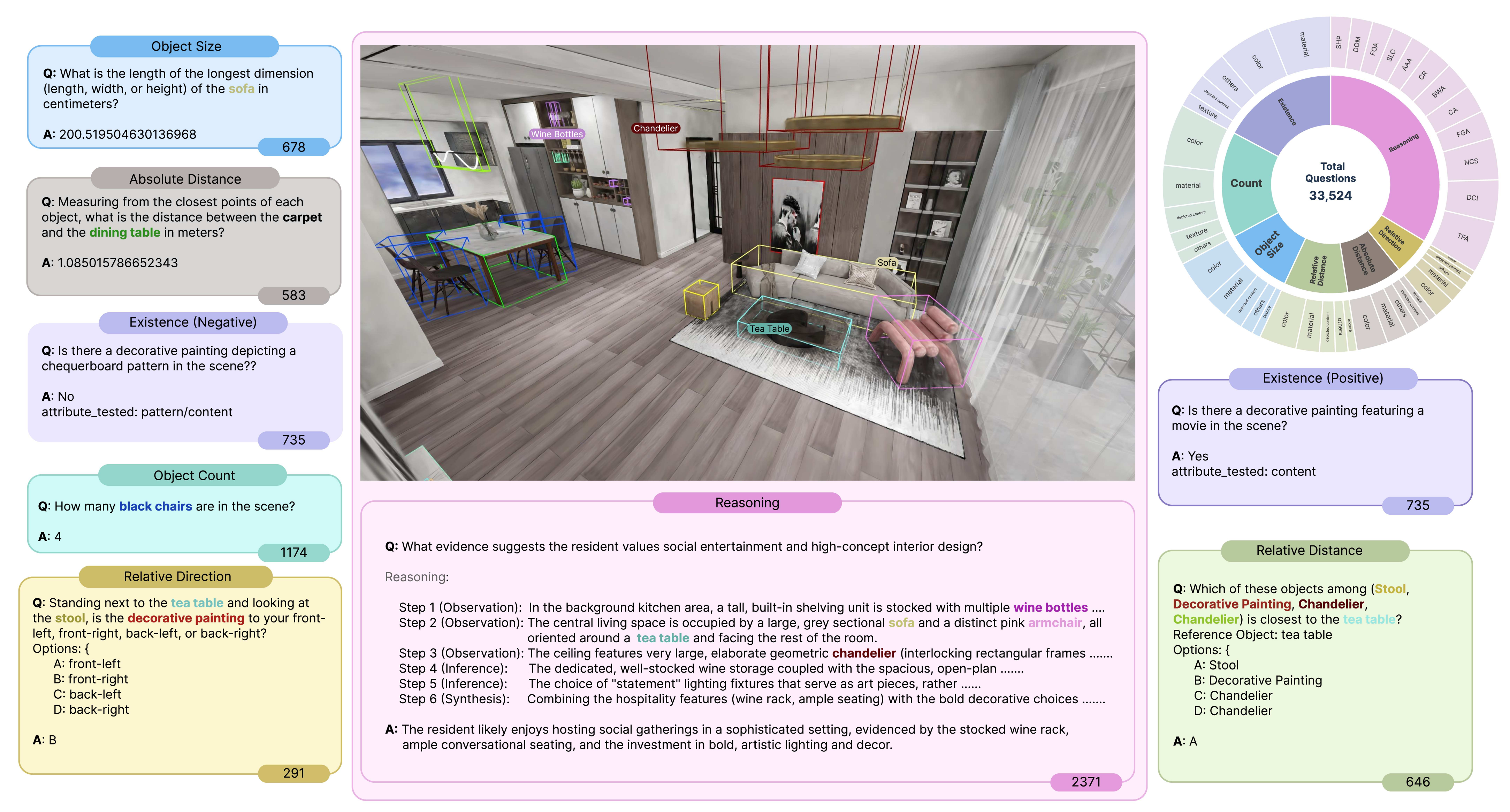}
    \caption{\textbf{(Center)} A 3D Gaussian Splatting scene with grounded object labels and bounding boxes embedded directly in the scene, referencing the objects involved in the answer to the \textit{Reasoning} question shown below it. \textbf{(Left and Bottom Right)} Representative examples of the remaining question types, each grounded to specific objects within the scene, with the numbers representing the total questions in the test set.
\textbf{(Top Right)} Distribution of the 33,524 total questions across the 7 major question types.}
    \label{fig:qa-section}
\end{figure*}

\subsection{Existence-Based Questions}
These questions ask whether a particular visual attribute of an object exists in the scene. They take the form of yes/no queries about properties such as color, material, or other visual characteristics.

To generate these questions, we first use the textual descriptions of objects in the scene, which contain detailed attributes. Given these descriptions, we prompt a vision-language model to generate both positive and negative questions for a given object category.

For positive (``yes") questions, the model selects an attribute that uniquely describes one instance of the object in the scene. For negative (``no") questions, it generates an attribute that does not appear in any instance of that object category within the scene.
Finally, all the generated questions for the test set are verified by the annotators.

\begin{table}[t]
\centering
\caption{Summary of Reasoning-Grounded Tasks and Definitions.}
\label{tab:qra_tasks}
\footnotesize
\begin{tabularx}{\linewidth}{l X}
\toprule
\textbf{Task} & \textbf{Core Definition} \\ \midrule
\textbf{NCS} & \textbf{Narrative and Contextual Synthesis:} Infers the "big picture" story, including user profiles and socioeconomic values reflected in the space. \\
\textbf{BWA} & \textbf{Behavior and Workflow Analysis:} Decodes patterns of action, such as daily habits or workflow efficiency in commercial spaces. \\
\textbf{DCI} & \textbf{Design and Curation Intent:} Analyzes deliberate aesthetic choices to understand the intended mood or brand identity. \\
\textbf{TFA} & \textbf{Temporal Forensic Analysis:} Focuses on recent activity by identifying transient evidence. \\
\textbf{FGA} & \textbf{Functional and Goal-Oriented Analysis:} Evaluates how effectively the layout achieves its primary purpose. \\
\textbf{CA} & \textbf{Comparative Analysis:} Compares two or more objects/areas to infer meaning from their differences. \\
\textbf{CR} & \textbf{Counterfactual Reasoning:} Tests the space's limitations by posing "what if" scenarios. \\
\textbf{AAA} & \textbf{Anomaly and Absence Analysis:} Identifies objects that are contextually out of place or missing. \\
\textbf{DOM} & \textbf{Dependent Object Manipulation:} Analyzes physical constraints and consequences of moving objects. \\
\textbf{SLC} & \textbf{Structural Load and Capacity:} Verifies if containment relationships are physically viable. \\
\textbf{FOA} & \textbf{Functional Occlusion and Accessibility:} Determines if object arrangements block the usage of a target. \\
\textbf{SHP} & \textbf{Systemic Hazard Propagation:} Traces potential damage paths in hypothetical failure scenarios. \\ \bottomrule
\end{tabularx}
\end{table}

\subsection{Spatial Intelligence Questions}
Inspired by~\cite{yang2025thinking}, we design questions that evaluate the model's spatial intelligence, covering object size, object count, absolute distance, relative distance, and relative direction. Ground truth for these tasks is derived directly from object metadata.

In contrast to~\cite{yang2025thinking}, our scenes contain multiple instances of objects from the same category. To address this complexity, we do not refer to objects by their class labels alone; instead, we utilize unique referring expressions based on visual attributes. These attributes encompass text, color, pattern, depicted content, material, and lighting effects. 

To generate referring expressions, we feed free-form descriptions and object images to a Vision-Language Model (VLM) to produce expressions that uniquely isolate the target within the scene. In the specific case of object counting, we employ a VLM to identify subsets of objects sharing a common attribute (e.g., "red shoes" rather than just "shoes"). This allows us to formulate precise counting queries that require the model to disentangle objects based on fine-grained visual features.
All the generated questions are verified by the annotators.

\subsection{Reasoning-Based Questions}
We generate complex reasoning questions that require a deep understanding of a scene's context, design, and implicit narratives. As illustrated in \Cref{tab:qra_tasks}, these tasks are classified into twelve distinct reasoning types. We utilize a vision-language model, prompted with both the scene video and our generated hierarchical descriptions, to produce one question per category for each scene where possible. Each entry is formatted as a Question-Reasoning-Answer (QRA) triplet, where the reasoning process justifies the final answer through a structured sequence of observations.

The reasoning steps are designed to mimic a human-like cognitive flow: starting with a Scene Overview to establish global context (Root level), followed by an Area Inspection to narrow focus to a specific zone (Child level), and Target Identification to pinpoint the object in question (Leaf level). To further refine the context, the model performs an Adjacent Observation of neighboring items (Sibling level) and a Fine Detail Observation of specific attributes such as text, brands, or material states. Finally, the model engages in Inference to apply deductive logic based on these observations, culminating in a Synthesis that combines all gathered evidence into a final conclusion.

\subsection{Quality and Statistics}
To ensure the quality of our generated questions and answers, we conducted a user study involving 100 questions across 10 distinct 3D scenes. Participants evaluated the accuracy and relevance of the content on a 5-point scale. To verify the reliability of these assessments, we included both clean samples and intentionally perturbed samples as a quality control measure.

As shown in \Cref{fig:flow}, users rated clean samples significantly higher than perturbed samples. More details can be found in the supplementary material.
    
\section{Experiments}
We evaluate the capacity of vision-language models for 3D scene understanding by benchmarking performance across three core dimensions: object existence, spatial intelligence, and complex reasoning. Our evaluation utilizes two distinct datasets---ScanNet++, representing real-world reconstructions, and InteriorGS, featuring high-fidelity synthetic Gaussian splats---to test the models across varied environmental complexities. Of the 966 scenes in total, we use 766 scenes (26,311 questions) for training and reserve the remaining 200 scenes (7,213 questions) for testing.

\begin{figure*}[t]
    \centering
    \includegraphics[width=\textwidth]{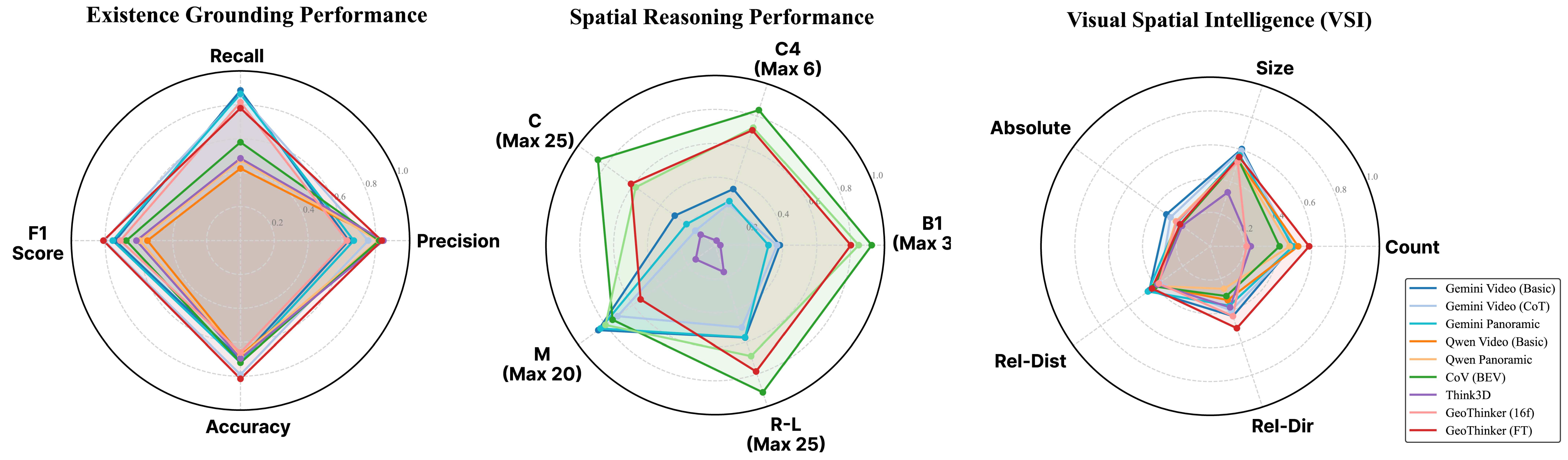}
\caption{\textbf{Granular performance breakdown.} Results of different models on specific performance categories: (a) existence grounding performance across precision, recall, F1, and accuracy; (b) reasoning performance across text-similarity metrics (BLEU, CIDEr, METEOR, ROUGE-L); (c) visual spatial intelligence across size, count, distance, and direction.}
    \label{fig:granular_existence_metrics}
\end{figure*}

\begin{table*}[t]
\centering
\caption{Existence and Spatial Intelligence benchmark results, reported side by side for ScanNet++ and InteriorGS. For Existence we report Precision (Prec), Recall (Rec), F1 Score, and Accuracy (Acc). For Spatial Intelligence we report Object Count (Cnt), Object Size (Sz), Absolute Distance (Abs), Relative Distance (Rel-D), and Relative Direction (Rel-Dir). The bold and underlined values represent the top-1 and top-2 performance among all models within each dataset and metric group, respectively.}
\label{tab:existence_spatial_categorized}
\footnotesize
\setlength{\tabcolsep}{2pt}
\begin{tabular}{l cccc ccccc c cccc ccccc}
\toprule
& \multicolumn{9}{c}{\textbf{ScanNet++}} & & \multicolumn{9}{c}{\textbf{InteriorGS}} \\
\cmidrule(lr){2-10} \cmidrule(lr){12-20}
& \multicolumn{4}{c}{Existence} & \multicolumn{5}{c}{Spatial Intelligence} & & \multicolumn{4}{c}{Existence} & \multicolumn{5}{c}{Spatial Intelligence} \\
\cmidrule(lr){2-5} \cmidrule(lr){6-10} \cmidrule(lr){12-15} \cmidrule(lr){16-20}
Method & Prec & Rec & F1 & Acc & Cnt & Sz & Abs & Rel-D & Rel-Dir & & Prec & Rec & F1 & Acc & Cnt & Sz & Abs & Rel-D & Rel-Dir \\
\midrule
\multicolumn{20}{l}{\textit{Gemini-Based Models}} \\
Video (Basic)          & 0.65 & 0.87 & 0.75 & 0.71 & 0.48 & \textbf{0.61} & 0.31 & \underline{0.48} & 0.48 & & 0.65 & \textbf{0.91} & 0.76 & 0.70 & 0.50 & \underline{0.60} & \textbf{0.34} & \underline{0.42} & 0.39 \\
Video (CoT)            & 0.76 & \underline{0.90} & \underline{0.82} & \underline{0.81} & 0.47 & \underline{0.60} & \underline{0.32} & 0.38 & 0.43 & & 0.76 & \underline{0.84} & \textbf{0.80} & \textbf{0.78} & \underline{0.54} & \textbf{0.64} & \underline{0.28} & 0.36 & \textbf{0.44} \\
Panoramic              & 0.66 & \textbf{0.95} & 0.78 & 0.74 & 0.45 & 0.57 & 0.27 & \textbf{0.52} & 0.43 & & 0.68 & 0.78 & 0.73 & 0.70 & 0.50 & 0.56 & 0.17 & 0.37 & 0.30 \\
\multicolumn{20}{l}{\textit{Qwen-Based Models}} \\
Video (Basic)          & \underline{0.85} & 0.56 & 0.67 & 0.73 & \underline{0.50} & 0.59 & 0.29 & 0.38 & 0.29 & & \underline{0.83} & 0.29 & 0.43 & 0.62 & \underline{0.54} & 0.48 & 0.21 & \underline{0.42} & 0.38 \\
Panoramic              & 0.82 & 0.54 & 0.65 & 0.71 & 0.47 & 0.57 & 0.24 & 0.37 & 0.23 & & 0.81 & 0.41 & 0.54 & 0.66 & 0.47 & 0.49 & 0.26 & \textbf{0.43} & 0.30 \\
\multicolumn{20}{l}{\textit{Agentic Methods}} \\
CoV                    & 0.82 & 0.59 & 0.68 & 0.73 & 0.44 & 0.56 & 0.21 & 0.41 & 0.35 & & 0.82 & 0.57 & 0.67 & \underline{0.71} & 0.38 & 0.52 & 0.22 & 0.40 & 0.27 \\
Think3D                & 0.83 & 0.54 & 0.66 & 0.72 & 0.23 & 0.34 & 0.20 & 0.38 & 0.37 & & \textbf{0.86} & 0.43 & 0.57 & 0.67 & 0.25 & 0.33 & 0.21 & 0.39 & 0.39 \\
\multicolumn{20}{l}{\textit{Geometry Integrated VLMs}} \\
GeoThinker (16 frames) & 0.60 & 0.86 & 0.71 & 0.65 & 0.19 & 0.57 & \textbf{0.33} & 0.47 & \underline{0.51} & & 0.66 & 0.77 & 0.71 & 0.67 & 0.24 & 0.48 & 0.16 & 0.28 & 0.36 \\
\multicolumn{20}{l}{\textit{Finetuned Model}} \\
GeoThinker (Finetuned) & \textbf{0.87} & 0.80 & \textbf{0.84} & \textbf{0.85} & \textbf{0.60} & 0.59 & 0.31 & \underline{0.48} & \textbf{0.62} & & 0.80 & 0.76 & \underline{0.78} & \textbf{0.78} & \textbf{0.57} & 0.52 & 0.13 & 0.37 & \underline{0.40} \\
\bottomrule
\end{tabular}
\end{table*}

\begin{table}[t]
\centering
\caption{Granular LLM-as-a-judge ratings using Qwen3-32B as the judge model, reported for InteriorGS (IntGS) and ScanNet++ (SNet++) side by side. The bold and underlined values represent the top-1 and top-2 performance among all models within each dataset, respectively.}
\label{tab:granular_ratings_final}
\footnotesize
\renewcommand{\arraystretch}{1.05}
\setlength{\tabcolsep}{2pt}
\begin{tabular}{l cccc cccc}
\toprule
& \multicolumn{4}{c}{\textbf{IntGS}} & \multicolumn{4}{c}{\textbf{SNet++}} \\
\cmidrule(lr){2-5} \cmidrule(lr){6-9}
& \multicolumn{3}{c}{Base} & FT & \multicolumn{3}{c}{Base} & FT \\
\cmidrule(lr){2-4} \cmidrule(lr){5-5} \cmidrule(lr){6-8} \cmidrule(lr){9-9}
Type & Vid.B & CoV & GeoT & GT-FT & Vid.B & CoV & GeoT & GT-FT \\
\midrule
AAA & \textbf{3.14} & 2.70 & 2.21 & \underline{2.79} & \textbf{3.27} & \underline{2.99} & 2.60 & 2.75 \\
BWA & \textbf{3.14} & 3.01 & 2.58 & \underline{3.07} & 2.60 & \textbf{3.17} & 2.82 & \underline{3.05} \\
CA  & \textbf{2.97} & \underline{2.82} & 2.54 & 2.79 & \textbf{3.47} & 2.91 & 2.43 & \underline{2.94} \\
CR  & \textbf{3.65} & \underline{2.92} & 2.53 & 2.82 & \textbf{3.24} & \underline{2.93} & 2.47 & 2.83 \\
DCI & \underline{3.29} & 2.96 & 2.84 & \textbf{3.30} & 2.76 & \textbf{3.43} & 3.08 & \underline{3.34} \\
DOM & \textbf{2.54} & 2.49 & 2.13 & \underline{2.51} & \textbf{3.21} & \underline{2.70} & 2.28 & 2.60 \\
FGA & \textbf{3.61} & \underline{3.38} & 2.88 & 3.22 & \underline{3.04} & 3.03 & 2.96 & \textbf{3.34} \\
FOA & \underline{2.99} & 2.92 & 2.92 & \textbf{3.08} & \textbf{3.02} & 2.71 & \underline{2.86} & 2.79 \\
NCS & 3.19 & \underline{3.49} & 3.11 & \textbf{3.62} & 2.45 & \textbf{3.35} & 2.99 & \underline{3.24} \\
SHP & \textbf{2.96} & 2.66 & 2.14 & \underline{2.72} & \textbf{3.17} & \underline{2.84} & 2.10 & 2.73 \\
SLC & \textbf{3.36} & 3.07 & 2.95 & \underline{3.08} & \textbf{3.94} & 3.25 & 3.45 & \underline{3.55} \\
TFA & \textbf{3.18} & 2.71 & 2.29 & \underline{2.93} & \underline{2.79} & \textbf{2.87} & 2.51 & 2.73 \\
\midrule
\textbf{Mean} & \textbf{3.17} & 2.93 & 2.59 & \underline{2.99} & \textbf{3.08} & \underline{3.01} & 2.71 & 2.99 \\
\bottomrule
\end{tabular}
\end{table}

\subsection{Evaluation Metrics}

We employ three sets of metrics corresponding to our three question categories:

\begin{itemize}
    \item \textbf{Existence Questions:} We evaluate performance using Precision, Recall, F1 Score, and Accuracy Score.

    \item \textbf{Spatial Intelligence Questions:} For spatial intelligence questions, we adopt the metrics from \textit{Thinking in Space}~\cite{yang2025thinking}. We use \textbf{Accuracy (ACC)} for Multiple-Choice Answer (MCA) tasks. For Numerical Answer (NA) tasks, which involve continuous value prediction, we use \textbf{Mean Relative Accuracy (MRA)}; this averages relative accuracy across confidence thresholds to measure proximity to the ground truth. 
    
    \item \textbf{Reasoning Questions:} For our Reasoning-Grounded QRA Triplets, we assess the quality of generated free-form answers against ground-truth responses. We report scores using standard n-gram and consensus-based language metrics: \textbf{BLEU-1}, \textbf{BLEU-4}, \textbf{CIDEr}~\cite{vedantam2015ciderconsensusbasedimagedescription}, \textbf{METEOR}~\cite{banerjee-lavie-2005-meteor}, and \textbf{ROUGE-L}~\cite{lin-2004-rouge}. Besides this, we also use Qwen3-32B as an LLM-as-a-judge to rate the model responses compared to the ground truth for Gemini on video inputs, CoV, GeoThinker and GeoThinker finetuned on our data; we use a different model family for judging, rather than Gemini, which was also used for question generation, to avoid self-preference bias.
\end{itemize}

\subsection{Methods}
We evaluate spatial reasoning across a diverse suite of baselines, including proprietary and open-source VLMs (e.g., ``gemini-3-flash-preview'', Qwen3-VL-8B~\cite{bai2025qwen3vltechnicalreport}) tested with various visual inputs like video or panoramas. Additionally, for Gemini-3-flash-preview we also test a prompting technique, CoT. These are compared against specialized 3D-aware frameworks: Chain-of-View (CoV)~\cite{zhao2026covchainofviewpromptingspatial}, which employs an agentic exploration of the 3D scene (represented as Gaussian splats in our evaluation); Think3D~\cite{zhang2026think3dthinkingspacespatial}, which utilizes an interactive 3D manipulation toolkit; and GeoThinker~\cite{li2026thinkinggeometryactivegeometry}, which features an active geometry integration mechanism. We also finetune GeoThinker on our dataset.

\subsection{Benchmark Results}

We present the evaluation results in this section, separated by dataset (ScanNet++ and InteriorGS).
InteriorGS consists of synthetic scenes featuring high-fidelity Gaussian splats, primarily focusing on large-scale apartments. In contrast, ScanNet++ is composed of reconstructions from real-world captures, typically representing smaller apartments and compact environments.

As illustrated in the Existence columns of \Cref{tab:existence_spatial_categorized}, Vision-Language Models (VLMs) utilizing video-based methods achieve impressive results in object existence tasks. Proprietary models, such as Gemini-3-flash, generally perform slightly better than open-source alternatives like Qwen-3-8B-VL. The application of Chain-of-Thought (CoT) reasoning further improves these results, allowing Video (CoT) to achieve the best performance across all primary metrics. Utilizing panoramic images as an alternative input yields slightly worse results than video input. Agentic methods, such as CoV and Think3D, follow as the next-best-performing category; notably, CoV outperforms Think3D in this category. This performance gap is likely due to CoV's ability to explore directly in high-fidelity Gaussian-splat environments, whereas Think3D relies on specialized 3D modules that may introduce noise during scene reconstruction. The disparity is particularly pronounced in synthetic scenes, where the quality of the underlying Gaussian splats provides a more reliable foundation for reasoning than the often fragmented or lower-resolution reconstructions found in real-world data. GeoThinker, which integrates geometric features, initially struggles on existence questions, but fine-tuning it on our data substantially increases its performance, as shown in \cref{fig:granular_existence_metrics}.

As illustrated in the Spatial Intelligence columns of \Cref{tab:existence_spatial_categorized}, VLMs demonstrate superior performance on counting-related tasks. This is likely because processing entire videos provides broader scene context; proprietary models like Gemini also consistently outperform open-source models like Qwen, except in some categories such as count. Consistent with Thinking in Space, applying Chain-of-Thought (CoT) reasoning tends to degrade performance on spatial intelligence tasks. For queries like relative distance, panoramic images often outperform the base Gemini model, particularly in smaller-scene datasets like ScanNet++, where all relevant objects fit within a single panoramic frame.

While video-based methods yield impressive results by leveraging extensive temporal context, they require a high volume of frames for evaluation, making them computationally heavy. In contrast, agentic methods that move around a scene do not perform well on spatial intelligence tasks, often struggling with spatial coherence as they transition between viewpoints. Conversely, methods that incorporate geometric features, such as GeoThinker, are better suited to fine-grained spatial queries like relative direction and distance, though the base model still struggles on others such as counting. Integrating a geometric encoder significantly boosts performance, and after fine-tuning on our data GeoThinker surpasses vision-language models that are provided with dense videos of the scene, as shown in \cref{fig:granular_existence_metrics}. This indicates that actively fusing geometric information into the architecture is a highly effective strategy, allowing a specialized, geometry-aware model to outperform general-purpose LLMs. 
More fine-grained results and analysis based on object attributes can be found in the supplementary material.

As reported in \Cref{tab:granular_ratings_final}, Gemini with dense visual input (Video (Basic)) is the best method overall, as its dense video sampling provides the most comprehensive scene context for reasoning. The main exception is Narrative and Contextual Synthesis (NCS), where CoV performs better, likely because its Bird's Eye View input provides an immediate, holistic blueprint of the room's layout. GeoThinker struggles with advanced reasoning, but fine-tuning it on our data increases its performance, as shown in \cref{fig:granular_existence_metrics} and \cref{tab:granular_ratings_final}. Text-based metrics have been provided in the supplementary material.

\section{Conclusion}
In this work, we introduce \textbf{SceneBench}, a novel large-scale benchmark of 966 photorealistic 3D scenes captured via Gaussian Splatting, designed to bridge the gap between 2D image analysis and true 3D spatial intelligence. 
By implementing a hierarchical, tree-based annotation structure spanning scenes, rooms, functional areas, and individual objects, we provide a benchmark that captures both fine-grained visual detail and the hierarchical structural organization of real-world environments. 
Our extensive evaluation of current vision-language models across existence, spatial intelligence, and multi-step reasoning tasks reveals significant performance bottlenecks, particularly as hierarchical complexity increases.
While models perform reasonably well on basic recognition tasks, their performance drops substantially when required to reason across semantic levels or integrate fine visual details with spatial relationships. 
These findings highlight limitations that are not exposed by existing benchmarks and demonstrate the need for new evaluation settings that reflect the complexity of real environments. As such, SceneBench provides a realistic testbed for diagnosing model capabilities and guiding the development of future vision-language systems with stronger spatial intelligence. We hope SceneBench will enable the community to develop and evaluate vision-language models with stronger spatial intelligence in photorealistic 3D environments.

{
    \small
    \bibliographystyle{ieeenat_fullname}
    \bibliography{main}
}

\clearpage
\setcounter{page}{1}
\maketitlesupplementary

\section{Overview}
\label{sec:Overview}
This supplementary includes additional materials not included in the main paper.

\begin{itemize}
    \item \textbf{\Cref{sec:Data Generation Process}:} Details the \textbf{Data Generation Process}, covering scene video generation, hierarchy generation, and question generation.

    \item \textbf{\Cref{sec:Data Statistics}:} Reports the \textbf{Data Statistics} of the released question set, broken down by question type, visual attribute, and reasoning sub-category.

    \item \textbf{\Cref{sec:Additional Results}:} Presents \textbf{Additional Results} not included in the main paper: reasoning text-similarity metrics, a granular per-attribute and per-hierarchy-level performance breakdown, results for GeoThinker with additional frames, and the average frame usage across methods.

    \item \textbf{\Cref{sec:Human Evaluation Interface}:} Provides an overview of the \textbf{Human Evaluation Interface} used for rating the quality of question-answer pairs.

    \item \textbf{\Cref{sec:Qualitative Results}:} Displays \textbf{Qualitative Results} illustrating the performance of various methods on our benchmarking tasks.
\end{itemize}

\section{Data Generation Process}
\label{sec:Data Generation Process}

\begin{figure*}[t]
    \centering
    \includegraphics[width=\textwidth]{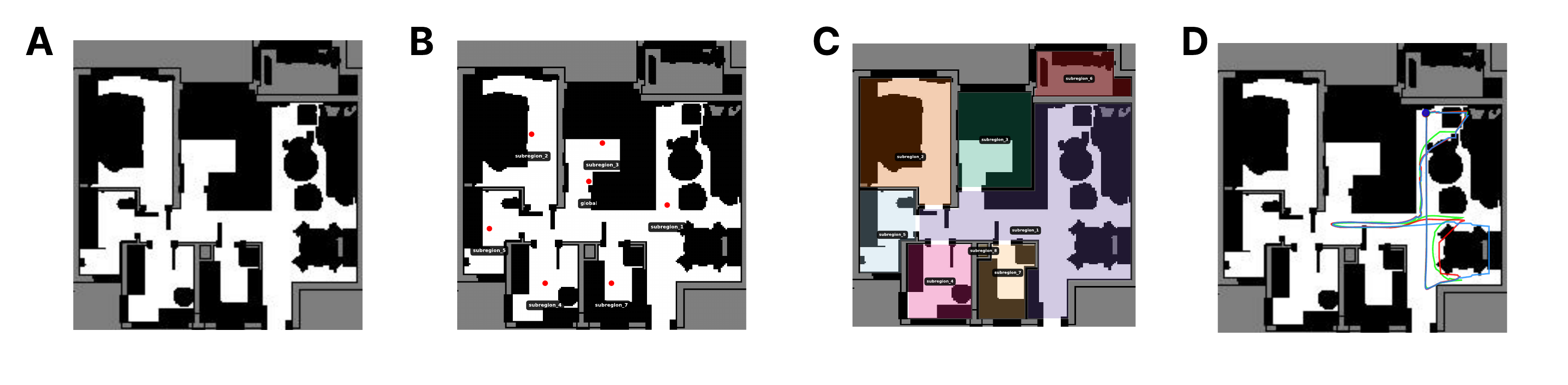}
    \caption{(A) The base 2D occupancy map provided by the InteriorGS dataset, serving as the foundational navigational grid. (B) Top-down view illustrating the sampled collision-free camera positions utilized for generating 360-degree panoramic images via 3D Gaussian Splatting. (C) The global scene cleanly partitioned into localized semantic subregions (e.g., individual rooms) using simplified polygon bounds to support targeted scene understanding. (D) Extracted 3D video camera trajectories plotted at low (floor), high (flying), and mid-level elevations. The elevation levels are color-coded as follows: lime (floor), red (mid-level), and dodgerblue (flying).}
    \label{fig:occupancy-maps}
\end{figure*}

\subsection{Video Generation Process}
For the ScanNet++ dataset~\cite{yeshwanth2023scannethighfidelitydataset3d}, video sequences are constructed by leveraging the high-quality continuous frames natively provided by the dataset.
In contrast, for the synthetic InteriorGS dataset~\cite{spatialverse2025interiorgs}, camera trajectories are programmatically generated by first processing the scene's 2D occupancy map to identify navigable spaces at three heights: floor, mid-level, and flying. To define the path, each subregion is partitioned into a $5 \times 5$ grid, from which the centroid of the largest connected navigable space in each cell is extracted as a waypoint. These waypoints are then linked using the A* pathfinding algorithm to create a raw, closed-loop navigation path that ensures comprehensive coverage of the interior space.

To transition these raw paths into smooth, natural motion, we apply B-spline interpolation combined with a curvature-aware sampling strategy. This method dynamically increases the density of points in high-curvature segments, effectively slowing the camera during sharp turns to mimic human-like observation. Finally, camera orientations are determined by calculating ``look-at'' vectors based on future path coordinates, which are smoothed via a moving average filter. These 2D trajectories are then projected into 3D space using height offsets from the occupancy metadata to generate the final world-to-camera matrices for rendering. On a sample of scenes we manually compared these automatically generated videos against videos recorded by moving a camera through the scene by hand, and found the two to be of comparable quality for annotation.

We overlay instance ids of the objects in the videos by projecting the center of the bounding box of the objects in the image. All the renderings for images are created using the gsplat~\cite{ye2024gsplatopensourcelibrarygaussian} rasterizer.

\subsection{Dataset Generation Process}
Every stage described below is human-in-the-loop: we inspect the model outputs, revise the prompts and model inputs in response to recurring failure cases, and re-run the stage until its output is satisfactory before moving on.

\subsubsection{Object Descriptions}
We pass the videos with instance ids overlaid on them, a file containing the list of objects where each object is represented by its class label, instance id, bounding box represented using center and extent, and a list that specifies how many instances of a particular object are present in the scene to a VLM. The prompt used for this process is shown in \cref{fig: description generation prompt}. The model outputs a JSON containing a list of objects where each object contains a class label, an instance id, the description, and the location of the object.
Then, we check whether the descriptions for all the objects are generated or not. For objects whose descriptions are missing, we sample images of those objects by rendering the objects from camera positions close to it.
Once we have the descriptions of all the objects, we use a validation prompt as shown in \cref{fig:  validation prompt}, where we send descriptions and the video to ask the VLM to correct any incorrect descriptions.

\subsubsection{Hierarchy Generation}
Our Hierarchy Generation is a multi-step process. The steps are designed to produce a spatially and semantically grounded scene hierarchy.

In the first pass of this process, we group the objects together based on their label names and proximity/location of their bounding boxes. For example, if many objects with label \textit{book} are close in the scene, they are grouped together in a \textit{books} label. Because objects that share a label can still be visually distinct, wherever possible an annotator manually splits such a group into separate sub-groups, for example separating a tray of chocolate croissants from a tray of strawberry tarts even though both instances carry the label \textit{pastry}. Additionally, we check for spatial relationships that can be derived from bounding box information. For example, if a \textit{wardrobe} object has \textit{cloth}, \textit{ornament} and other such objects inside of it, they all will be grouped in a \textit{wardrobe and its parts} label. This process continues until all objects are associated with one of the 3 parent classes: Floor, Wall, or Ceiling.

On the second pass, we take this algorithmically generated hierarchy, along with the object description json and video of the subregion and feed it to a VLM with the hierarchy generation prompt as shown in \cref{fig: Hierarchy generation prompt}. This then gives us the hierarchy structure for each of the divided subregion.

We then join these hierarchies under a same parent node, and the description and label of this node is generated by providing the VLM with the scene video and the description of the immediate child objects of the root node. The prompt for generation of scene description is shown in \cref{fig: Scene Description generation prompt}.

Finally, to ensure proper grounded descriptions, we query whether each inner node has the mention of its immediate children or not and generate a list of ungrounded objects and its parent description that should contain them. Then, we provide the VLM with this information along with the subregion video to ensure proper grounded description at every level of the hierarchy. The prompt used for ensuring grounding in descriptions is shown in \cref{fig: Grounding Correction prompt}.

\subsubsection{Existence Based Questions}
For each unique object label in the scene (e.g., \textit{cushion}, \textit{painting}), all instances sharing that label are first aggregated into a single group, and two questions are generated per group, one positive and one negative. The positive question targets an attribute present in at least one instance of the group, while the negative question targets a plausible distractor attribute that is physically possible for the object category but visually absent from every instance in the group. This visual dependency constraint ensures that negative questions cannot be answered by common sense alone and require genuine visual inspection. As shown in \cref{fig: existence question generation prompt}, only attributes that require high-fidelity rendering to perceive are considered, while attributes such as shape, size, or object state are explicitly excluded. The annotators subsequently check all questions in the test set and correct them if needed.

\subsubsection{Spatial Intelligence Based Questions}
From the object descriptions, we use a label based filter to select objects that usually have high detailed visual features in them like \textit{paintings} etc. After this we search for high detailed visual features key-words in the description of the objects to select additional objects. If there are multiple instances in a class then all instances are selected in this process. If there are multiple instances of an object class, then we render some representative images for each object.
These images and the object descriptions are sent to a vision language model along with the prompt shown in \cref{fig: referring expression generation prompt} to generate unique referring expressions for each object in the class. Long free-form descriptions are converted into short, distinct referring expressions to refer to the objects. Additionally, we also extract the common features from the objects in this class during this step. The common features are used to create object count questions. We use the referring expressions together with the object geometry to generate the rest of the spatial intelligence questions (object size, absolute distance, relative distance, and relative direction). The ground-truth metric quantities for these questions (object dimensions and inter-object distances) are computed directly from the 3D object bounding boxes provided with the original datasets, ScanNet++~\cite{yeshwanth2023scannethighfidelitydataset3d} and InteriorGS~\cite{spatialverse2025interiorgs}. The questions are created using the same template as in this work~\cite{yang2025thinking}. The annotators subsequently check all questions in the test set and correct them if needed.

\subsubsection{Reasoning Based Questions}
Reasoning-based questions test deep spatial, functional, and causal understanding of the scene, going beyond simple attribute or count queries. As shown in \cref{fig: reasoning question generation prompt}, the VLM is provided with the scene video and the full scene hierarchy JSON, and is tasked with generating at least one question from each of 12 distinct reasoning categories. These categories span a broad spectrum of inferential difficulty, ranging from high-level narrative synthesis to fine-grained physical reasoning.

To ensure that answers are grounded in observable evidence rather than free-form speculation, each question is accompanied by a structured reasoning chain that follows a strict observational protocol, progressing from a broad scene overview down to specific object-level details, before arriving at an inference and final synthesis. Object instance IDs from the scene hierarchy are explicitly referenced at each step to anchor the reasoning in the scene's spatial structure.

\subsubsection{Question Attribute and Level}
\label{sec:question attribute and level}
For each existence question we record which visual feature of the referenced object class the question probes; this may itself be more than one feature (e.g.\ \emph{color and material}). For spatial intelligence questions the situation is richer: every referring expression carries its own list of features, and a question is built from several referring expressions, so it is associated with the features of multiple objects. This attribute mapping drives the statistics in \cref{tab:existence_attribute_statistics,tab:vsi_attribute_statistics} and the fine-grained comparison in \cref{fig:granular_comparison}. Questions probing multiple attributes are counted redundantly across all relevant attribute categories for both existence and visual-spatial intelligence tasks.
Furthermore, from the hierarchical descriptions of the objects in the scene, we can determine the depth level associated with each object. We use the maximum level over all the objects associated with a question as the level of that question. We present the level-based breakdown for existence and visual-spatial intelligence questions in \cref{fig:granular_comparison}.

\section{Data Statistics}
\label{sec:Data Statistics}

\begin{table}[t]
\centering
\caption{Question counts by type, split into the training and held-out test sets. \textit{Reasoning} aggregates the twelve sub-categories detailed in \cref{tab:reasoning_statistics}.}
\label{tab:dataset_statistics}
\footnotesize
\setlength{\tabcolsep}{8pt}
\begin{tabular}{lrrr}
\toprule
\textbf{Question type} & \textbf{Test} & \textbf{Train} & \textbf{Total} \\
\midrule
Existence           & 1{,}470 & 4{,}300 &  5{,}770 \\
Count               & 1{,}174 & 4{,}073 &  5{,}247 \\
Object Size         &   678 & 2{,}750 &  3{,}428 \\
Absolute Distance   &   583 & 2{,}187 &  2{,}770 \\
Relative Distance   &   646 & 2{,}516 &  3{,}162 \\
Relative Direction  &   291 & 1{,}577 &  1{,}868 \\
Reasoning           & 2{,}371 & 8{,}908 & 11{,}279 \\
\midrule
\textbf{All types}  & \textbf{7{,}213} & \textbf{26{,}311} & \textbf{33{,}524} \\
\bottomrule
\end{tabular}
\end{table}

\begin{table}[t]
\centering
\caption{Granular breakdown of the \textit{Reasoning} questions by sub-category (Test\,/\,Train\,/\,Total counts); the corresponding task definitions are given in the main paper. The twelve categories are sampled near-uniformly, one question per category per scene wherever possible.}
\label{tab:reasoning_statistics}
\footnotesize
\setlength{\tabcolsep}{5pt}
\resizebox{\linewidth}{!}{
\begin{tabular}{llrrr}
\toprule
& \textbf{Sub-category} & \textbf{Test} & \textbf{Train} & \textbf{Total} \\
\midrule
NCS & Narrative and Contextual Synthesis     & 203 & 743 & 946 \\
BWA & Behavior and Workflow Analysis         & 195 & 741 & 936 \\
DCI & Design and Curation Intent             & 203 & 743 & 946 \\
TFA & Temporal Forensic Analysis             & 208 & 744 & 952 \\
FGA & Functional and Goal-Oriented Analysis  & 195 & 743 & 938 \\
CA  & Comparative Analysis                   & 195 & 742 & 937 \\
CR  & Counterfactual Reasoning               & 195 & 742 & 937 \\
AAA & Anomaly and Absence Analysis           & 197 & 742 & 939 \\
DOM & Dependent Object Manipulation          & 195 & 742 & 937 \\
SLC & Structural Load and Capacity           & 195 & 742 & 937 \\
FOA & Functional Occlusion and Accessibility & 195 & 742 & 937 \\
SHP & Systemic Hazard Propagation            & 195 & 742 & 937 \\
\midrule
\multicolumn{2}{l}{\textbf{All reasoning}}   & \textbf{2{,}371} & \textbf{8{,}908} & \textbf{11{,}279} \\
\bottomrule
\end{tabular}
}
\end{table}

\begin{table}[t]
\centering
\caption{\textbf{Visual-feature composition of the Existence questions} (Test\,/\,Train).}
\label{tab:existence_attribute_statistics}
\footnotesize
\setlength{\tabcolsep}{8pt}
\begin{tabular}{lrr}
\toprule
\textbf{Feature bucket} & \textbf{Test} & \textbf{Train} \\
\midrule
color               & 496 & 1{,}626 \\
material            & 645 & 2{,}286 \\
texture            &  81 &   270 \\
text               &  92 &    82 \\
depicted content   & 259 &   221 \\
lighting properties &  29 &    84 \\
\midrule
\textit{questions}  & \textit{1{,}470} & \textit{4{,}300} \\
\bottomrule
\end{tabular}
\end{table}

\begin{table*}[t]
\centering
\caption{\textbf{Feature composition of the visual--spatial intelligence question types} (Test\,/\,Train).}
\label{tab:vsi_attribute_statistics}
\footnotesize
\setlength{\tabcolsep}{4pt}
\begin{tabular}{l cc cc cc cc cc}
\toprule
& \multicolumn{2}{c}{\textbf{Count}} & \multicolumn{2}{c}{\textbf{Obj.\ Size}} & \multicolumn{2}{c}{\textbf{Abs.\ Dist.}} & \multicolumn{2}{c}{\textbf{Rel.\ Dist.}} & \multicolumn{2}{c}{\textbf{Rel.\ Dir.}} \\
\cmidrule(lr){2-3}\cmidrule(lr){4-5}\cmidrule(lr){6-7}\cmidrule(lr){8-9}\cmidrule(lr){10-11}
Feature bucket & Test & Train & Test & Train & Test & Train & Test & Train & Test & Train \\
\midrule
color               & 771 & 2{,}994 & 543 & 2{,}251 & 553 & 2{,}062 & 646 & 2{,}514 & 286 & 1{,}532 \\
material            & 497 & 1{,}862 & 260 & 1{,}035 & 347 & 1{,}249 & 571 & 2{,}151 & 230 & 1{,}120 \\
texture            & 314 & 1{,}176 & 121 &   467 & 178 &   619 & 345 & 1{,}220 & 149 &   629 \\
text               & 361 & 1{,}327 & 147 &   560 & 214 &   750 & 406 & 1{,}471 & 164 &   728 \\
depicted content   & 365 & 1{,}544 & 182 &   735 & 268 &   984 & 462 & 1{,}805 & 139 &   980 \\
lighting properties &  21 &    92 &   6 &    28 &  18 &    55 &  29 &   107 &   8 &    56 \\
Other              & 223 &   615 & 112 &   520 & 218 &   773 & 343 & 1{,}409 & 104 &   816 \\
\midrule
\textit{questions}  & \textit{1{,}174} & \textit{4{,}073} & \textit{678} & \textit{2{,}750} & \textit{583} & \textit{2{,}187} & \textit{646} & \textit{2{,}516} & \textit{291} & \textit{1{,}577} \\
\bottomrule
\end{tabular}
\end{table*}

\Cref{tab:dataset_statistics} summarizes the composition of the benchmark. The dataset contains 33{,}524 question--answer pairs (7{,}213 test and 26{,}311 train) drawn from 966 scenes: 584 real-world scenes from ScanNet++~\cite{yeshwanth2023scannethighfidelitydataset3d} and 382 synthetic scenes from InteriorGS~\cite{spatialverse2025interiorgs}, split into 766 training and 200 held-out test scenes. Reasoning is the largest category (11{,}279 questions); its per-sub-category counts are given in \cref{tab:reasoning_statistics}, where the twelve categories are seen to be sampled near-uniformly, one question per category per scene wherever possible.

\Cref{tab:existence_attribute_statistics,tab:vsi_attribute_statistics} report which features are used to pick out the objects each question refers to. A question that names two features (e.g.\ a \emph{white marble vase}, referenced by color and material) is counted under both, so a question can appear in more than one row and a column need not sum to the question total.

Each Existence question probes a single visual feature of the referenced object class. Color and material are more frequent compared to texture, depicted content, text, and lighting properties attributes.
Visual--Spatial Intelligence questions contain multiple objects referred to by different visual attributes, and attributes besides color and material are also well represented. Entries in the \emph{Other} category primarily contain non-visual properties, such as object positions.

\section{Additional Results}
\label{sec:Additional Results}

\subsection{Reasoning Text-Similarity Metrics}
\label{sec:reasoning text metrics}
\Cref{tab:reasoning_categorized} reports the n-gram and consensus-based text-similarity metrics for the free-form reasoning answers, complementing the LLM-as-a-judge ratings reported in the main paper. Across both datasets, the agentic method CoV and the fine-tuned GeoThinker produce answers that most closely match the ground-truth phrasing, with GeoThinker (Finetuned) achieving the highest BLEU-1, CIDEr, METEOR, and ROUGE-L scores on both ScanNet++ and InteriorGS. 
Conversely, the dense-video VLM (Gemini) scores lower on text-based metrics despite being the top-performing model under the LLM-as-a-judge evaluation. This discrepancy is likely due to Gemini's responses being significantly longer because they contain more detailed step-by-step reasoning than those of other models (\Cref{tab:response_length}), which is penalized by n-gram matching scores.

\begin{table}[t]
\centering
\caption{Mean answer length (words) of the LLM-judged models, with the ground-truth reasoning answers for reference. Computed over the $2{,}340$ test reasoning triplets.}
\label{tab:response_length}
\footnotesize
\setlength{\tabcolsep}{10pt}
\begin{tabular}{lrr}
\toprule
 & \textbf{Mean} & \textbf{Median} \\
\midrule
Ground-truth answer      &  31.6 & 31 \\
Gemini Video (Basic)     & 129.7 & 88 \\
CoV                      &  37.3 & 36 \\
GeoThinker (16 frames)   &  29.9 & 23 \\
GeoThinker (Finetuned)   &  28.7 & 29 \\
\bottomrule
\end{tabular}
\end{table}

\begin{table}[t]
\centering
\caption{Reasoning benchmark results. We report BLEU-1 (B1), BLEU-4 (B4), CIDEr (C), METEOR (M), and ROUGE-L (R-L). The bold and underlined values represent the top-1 and top-2 performance among all models, respectively.}
\label{tab:reasoning_categorized}
\resizebox{\linewidth}{!}{
\begin{tabular}{l ccccc c ccccc}
\toprule
& \multicolumn{5}{c}{\textbf{ScanNet++}} & & \multicolumn{5}{c}{\textbf{InteriorGS}} \\
\cmidrule(lr){2-6} \cmidrule(lr){8-12}
Method & B1 & B4 & C & M & R-L & & B1 & B4 & C & M & R-L \\
\midrule
\multicolumn{12}{l}{\textit{Gemini-Based Models}} \\
Video (Basic)          & 10.32 & 1.83 & 6.79 & 13.58 & 16.03 & & 12.45 & 2.33 & 7.88 & \underline{15.19} & 18.15 \\
Video (CoT)            & 11.16 & 1.55 & 3.03 & 11.81 & 12.53 & & 10.74 & 1.50 & 4.06 & 13.71 & 15.96 \\
Panoramic              &  8.92 & 1.60 & 3.58 & 14.03 & 16.01 & &  9.80 & 1.63 & 6.41 & 14.51 & 17.29 \\
\multicolumn{12}{l}{\textit{Qwen-Based Models}} \\
Video (Basic)          &  6.88 & 1.32 & 0.00 & 12.38 & 11.69 & &  7.48 & 1.35 & 0.00 & 12.57 & 12.62 \\
Panoramic              &  5.56 & 1.08 & 0.00 & 10.98 &  9.22 & &  5.34 & 1.01 & 0.00 & 10.85 &  8.91 \\
\multicolumn{12}{l}{\textit{Agentic Methods}} \\
CoV                    & \underline{28.27} & \underline{5.24} & \underline{23.00} & \underline{15.14} & \underline{23.10} & & \underline{27.21} & \textbf{4.80} & \underline{19.82} & 14.83 & \underline{22.54} \\
Think3D                &  1.38 & 0.25 &  2.58 &  2.92 &  4.05 & &  0.47 & 0.09 &  2.69 &  2.76 &  4.26 \\
\multicolumn{12}{l}{\textit{Geometry Integrated VLMs}} \\
GeoThinker (16 frames) & 22.12 & 3.88 & 14.27 & 10.25 & 18.74 & & 25.96 & \underline{4.67} & 16.46 & 11.53 & 20.42 \\
GeoThinker (Finetuned) & \textbf{35.52} & \textbf{9.81} & \textbf{58.53} & \textbf{16.19} & \textbf{30.44} & & \textbf{36.01} & 0.97 & \textbf{61.73} & \textbf{16.52} & \textbf{31.52} \\
\bottomrule
\end{tabular}
}
\end{table}

\subsection{Granular Performance Breakdown}
\label{sec:granular breakdown}
\begin{figure*}[t]
    \centering
    \includegraphics[width=\textwidth]{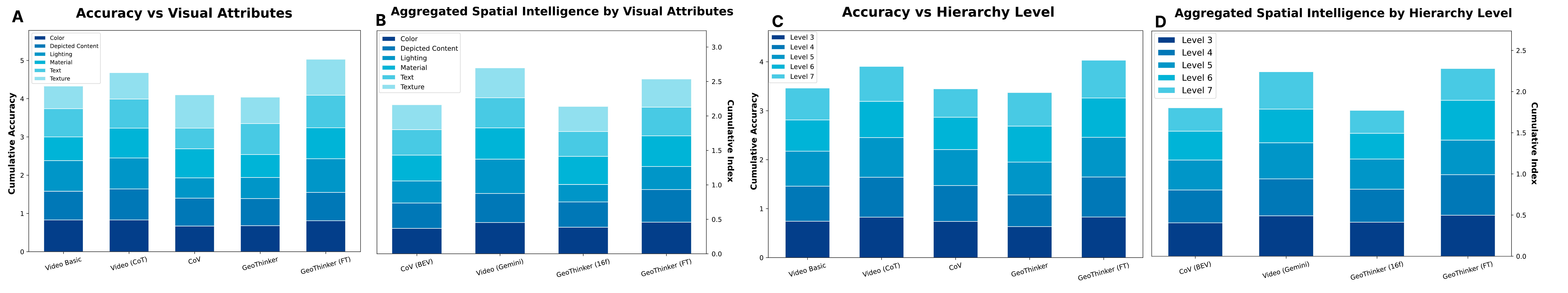}
    \caption{\textbf{Granular breakdown of existence and visual--spatial intelligence performance.} (a) Object-existence accuracy categorized by the probed visual attribute; (b) visual--spatial intelligence performance relative to the same attributes; (c) existence accuracy across hierarchy levels, from global scene level (Root) to fine-grained object level (Leaf); (d) visual--spatial intelligence consistency across hierarchy levels.}
    \label{fig:granular_comparison}
\end{figure*}

\begin{figure*}[t]
    \centering
    \includegraphics[width=\textwidth]{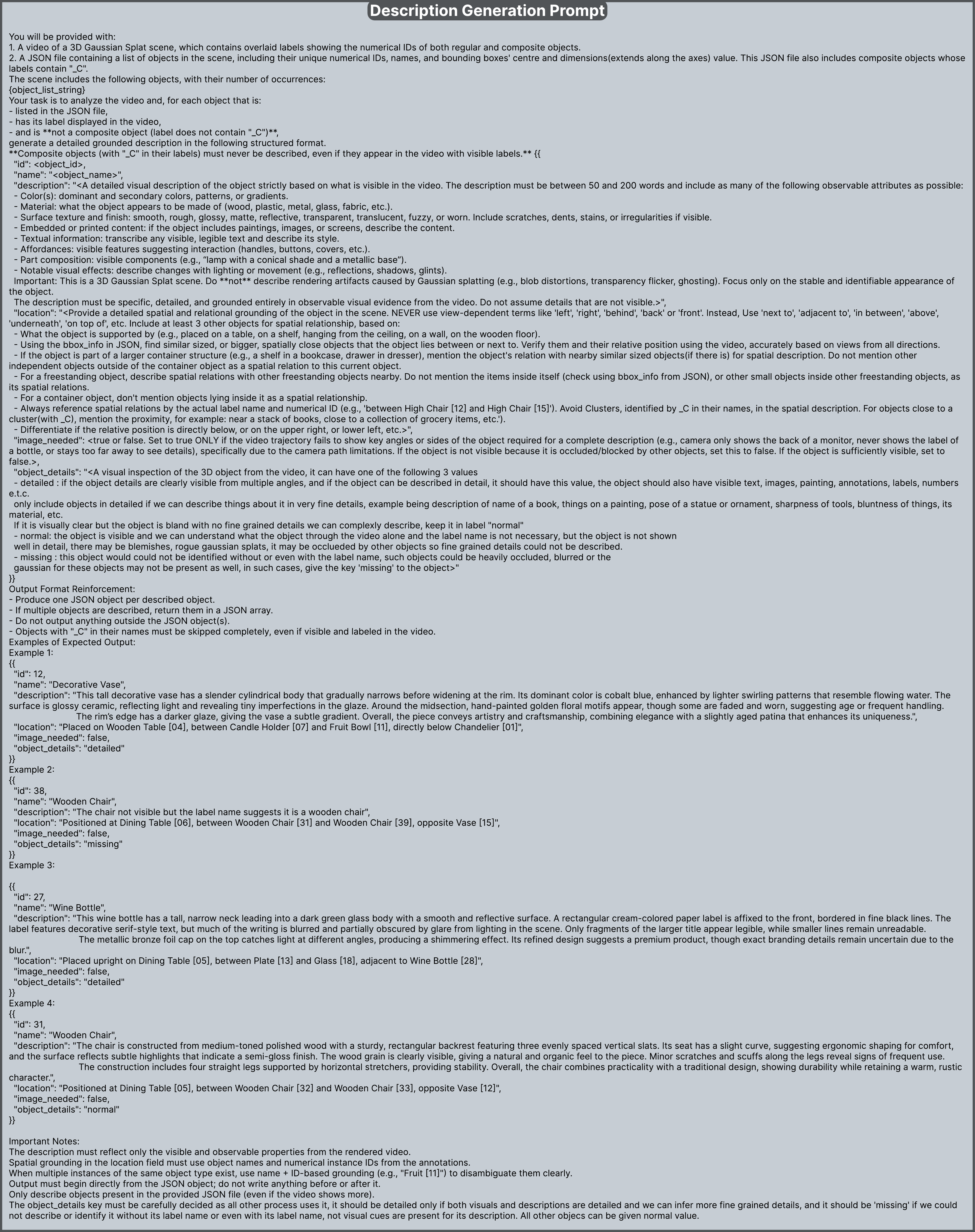}
    \caption{Description Generation Prompt.}
    \label{fig: description generation prompt}
\end{figure*}

\begin{figure*}[t]
    \centering
    \includegraphics[width=\textwidth]{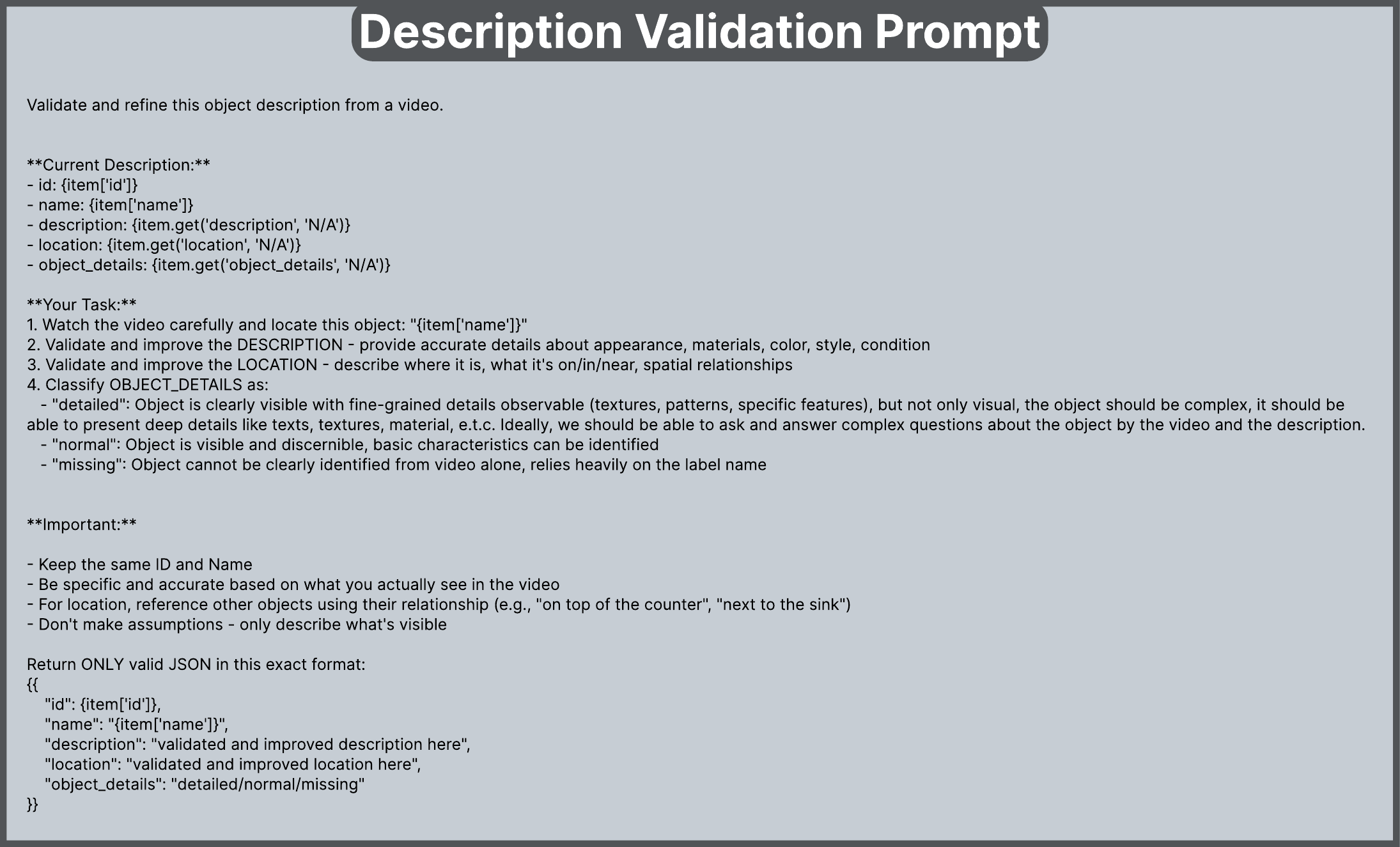}
    \caption{Validation Prompt.}
    \label{fig:  validation prompt}
\end{figure*}

\begin{figure*}[t]
    \centering
    \includegraphics[width=\textwidth]{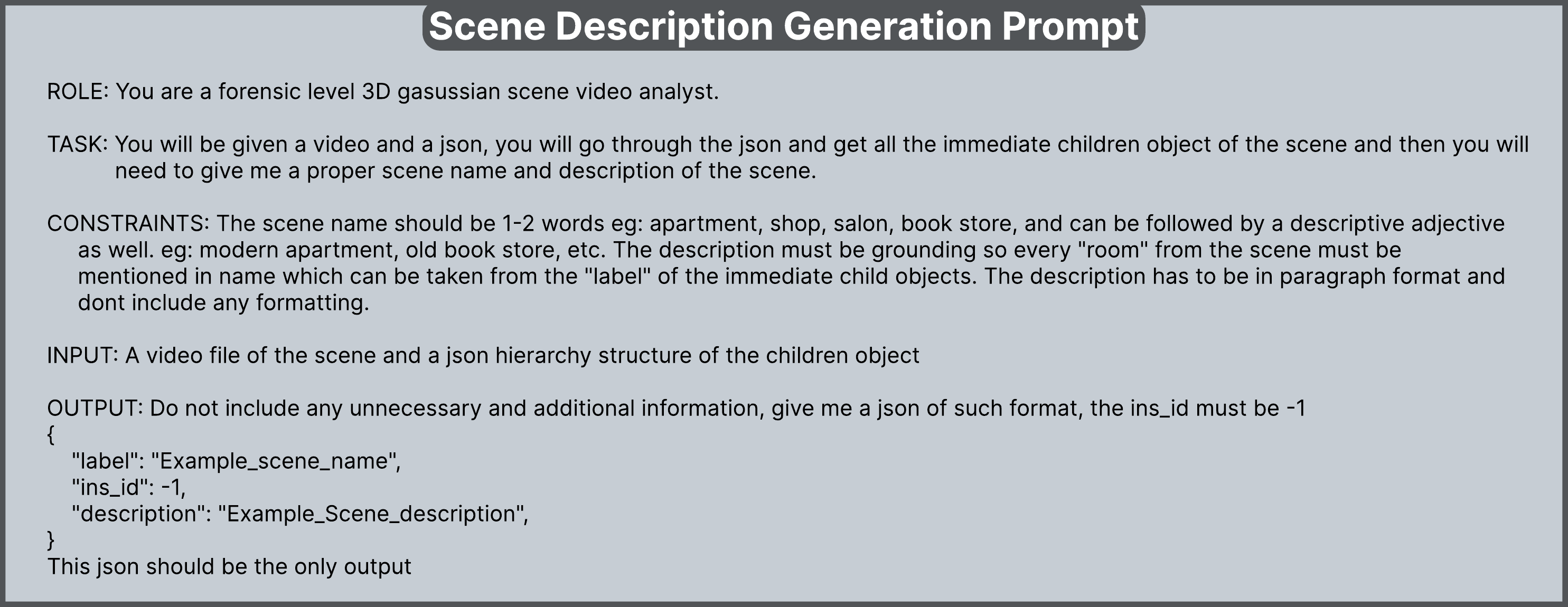}
    \caption{Scene Description Generation Prompt.}
    \label{fig: Scene Description generation prompt}
\end{figure*}

\begin{figure*}[t]
    \centering
    \includegraphics[width=\textwidth,height=0.95\textheight,keepaspectratio]{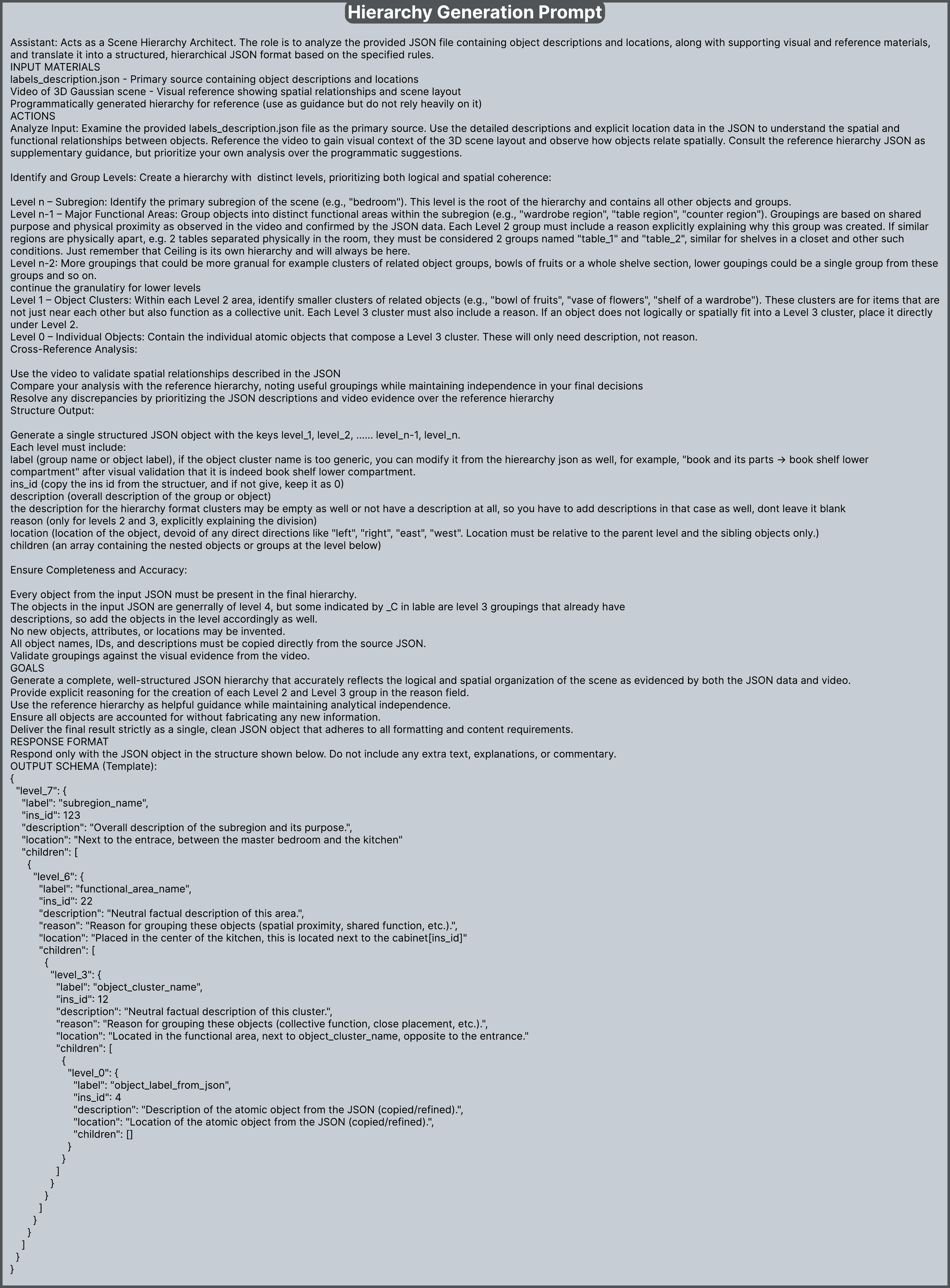}
    \caption{Hierarchy Generation Prompt.}
    \label{fig: Hierarchy generation prompt}
\end{figure*}

\begin{figure*}[t]
    \centering
    \includegraphics[width=\textwidth]{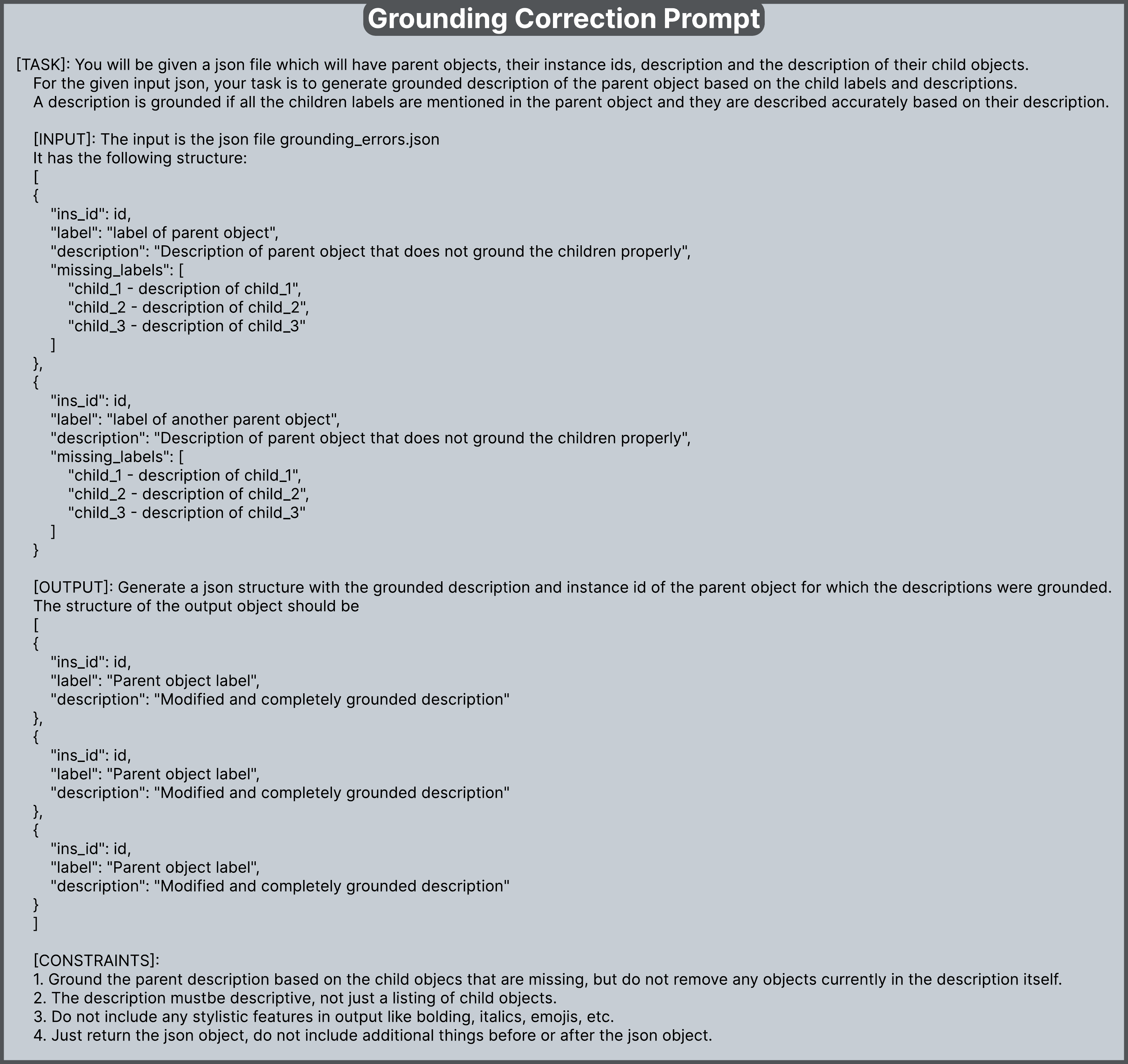}
    \caption{Grounding Correction Prompt.}
    \label{fig: Grounding Correction prompt}
\end{figure*}

\begin{figure*}[t]
    \centering
    \includegraphics[width=\textwidth,height=0.95\textheight,keepaspectratio]{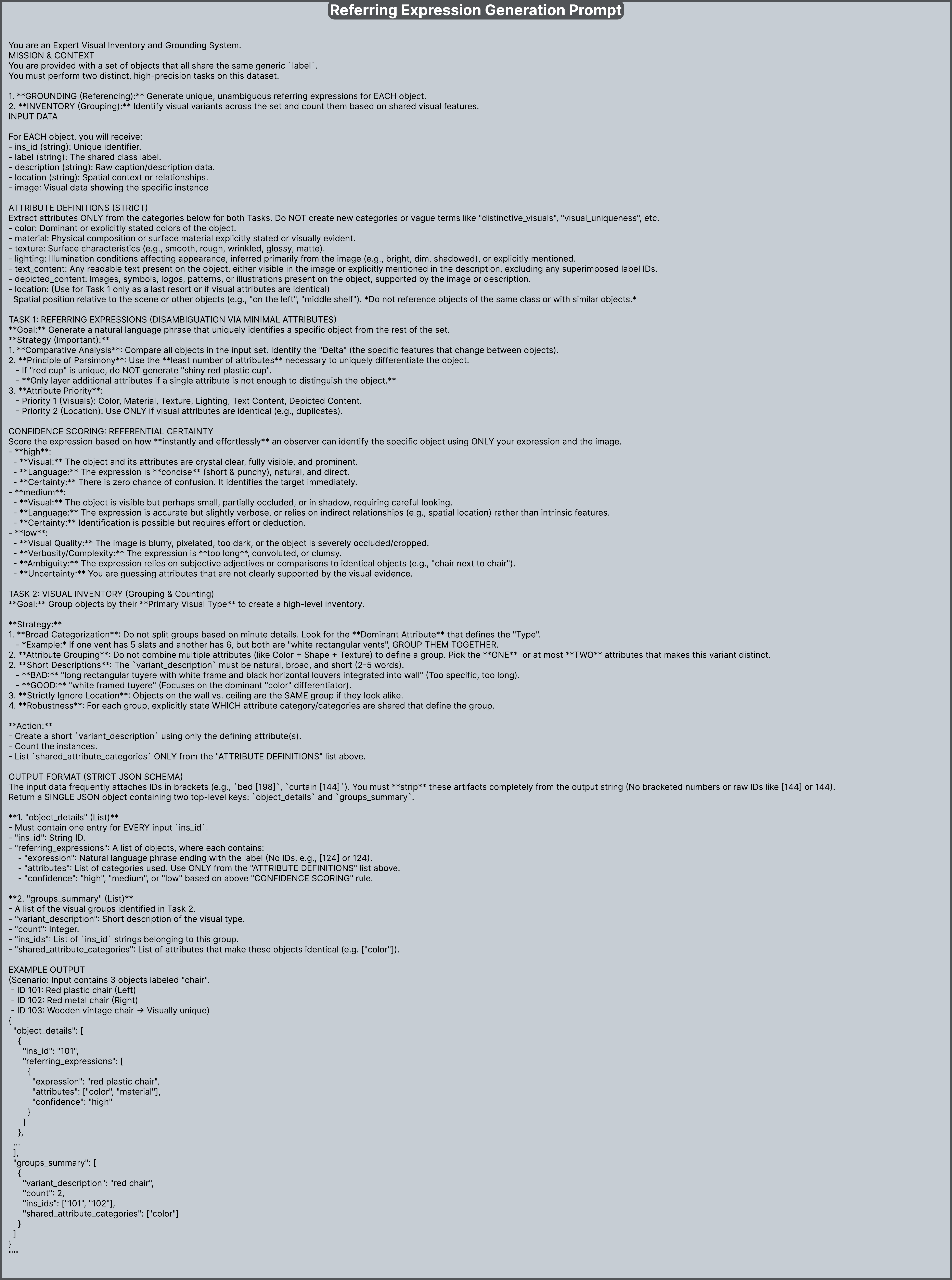}
    \caption{Referring Expression Generation Prompt.}
    \label{fig: referring expression generation prompt}
\end{figure*}

\begin{figure*}[t]
    \centering
    \includegraphics[width=\textwidth]{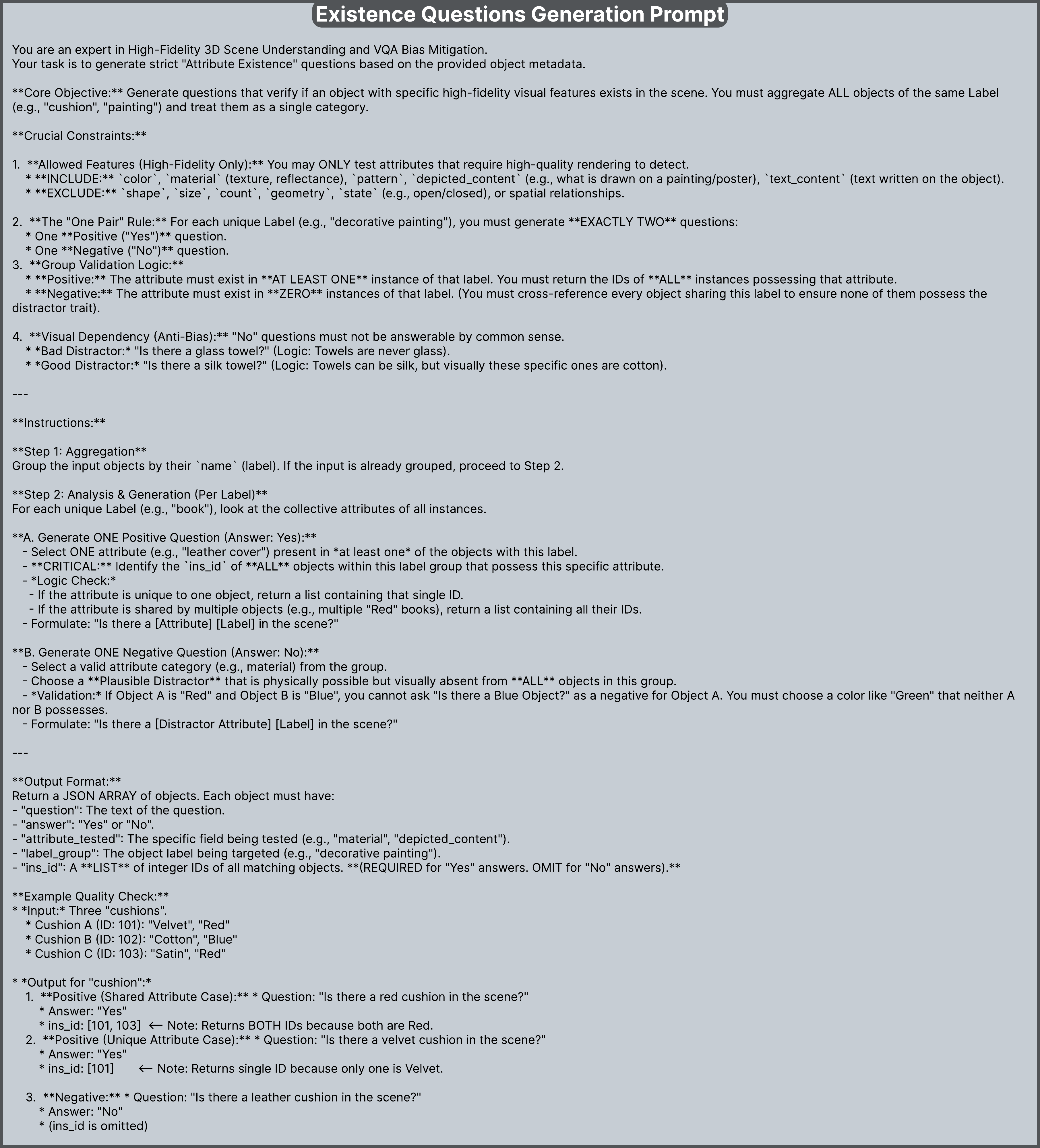}
    \caption{Existence Question Generation Prompt.}
    \label{fig: existence question generation prompt}
\end{figure*}

\begin{figure*}[t]
    \centering
    \includegraphics[width=\textwidth]{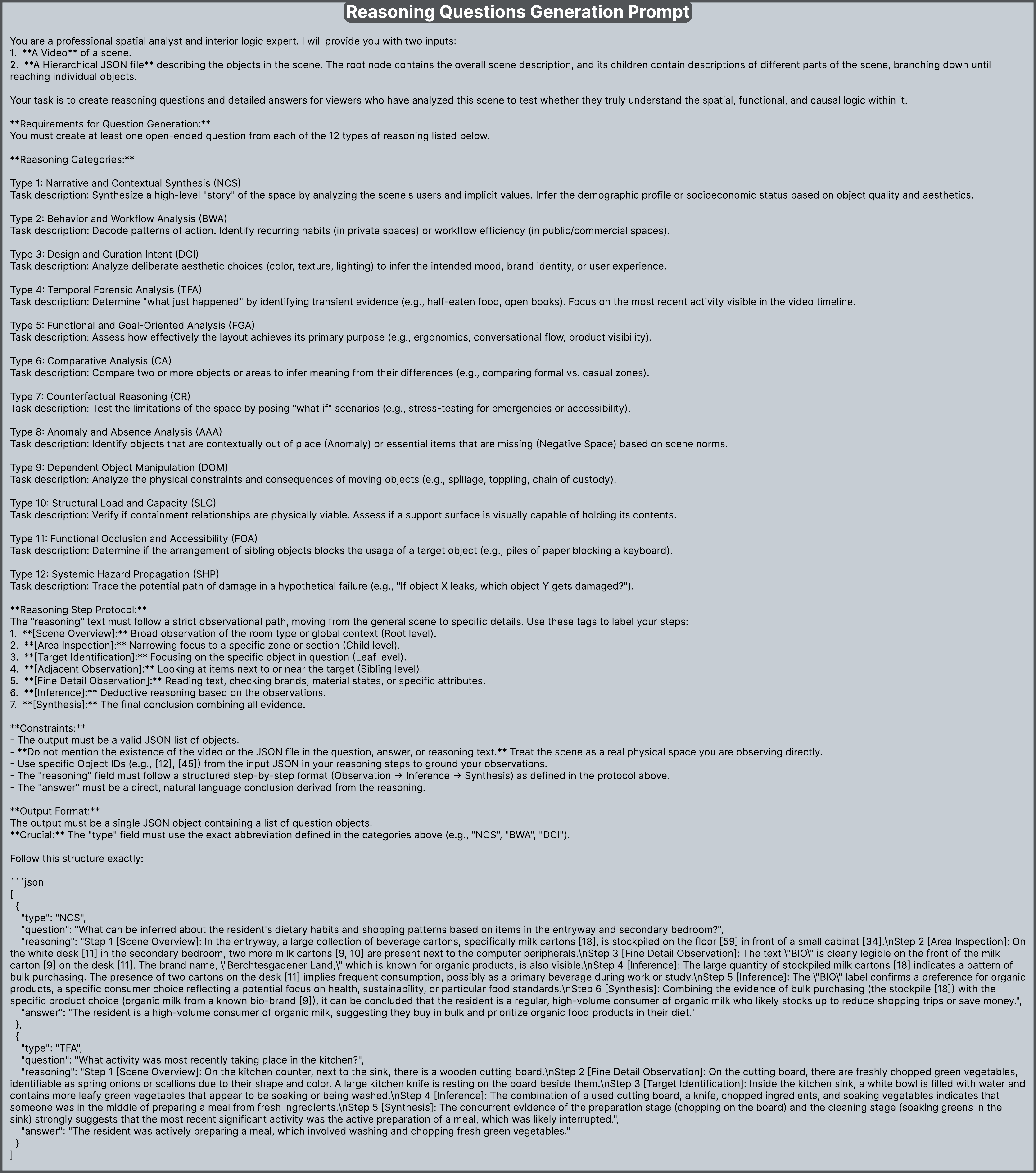}
    \caption{Reasoning Question Generation Prompt.}
    \label{fig: reasoning question generation prompt}
\end{figure*}

\Cref{fig:granular_comparison} presents model performance along two axes: the visual attribute a question probes, and the depth of the referenced object in the scene hierarchy. In visual-attribute recognition, video-based VLMs are consistent across all attribute categories, indicating a robust ability to perceive fine-grained visual detail; GeoThinker reaches a comparable profile only after fine-tuning on our data. Along the hierarchy-depth axis, most models perform well up to levels~4--5 and struggle at higher depths. Video (CoT) and GeoThinker (Finetuned) are the only methods that retain accuracy at the deepest (leaf) levels.

\subsection{Results of GeoThinker with more frames}
As illustrated in \cref{tab:existence_geothinker_more_frames}, \cref{tab:spatial_geothinker_more_frames} and \cref{tab:reasoning_geothinker_more_frames}, increasing the number of image frames improves the performance of GeoThinker~\cite{li2026thinkinggeometryactivegeometry} for most tasks.

\subsection{Frame Usage Across Methods}
\Cref{tab:frames_comparison} reports the average number of frames
consumed per question for all evaluated methods. Non-agentic methods such as Video (Basic) consume a fixed frame budget determined entirely by the video length, resulting in substantially higher frame counts (165--330 frames on ScanNet++). In contrast, agentic and geometry-aware methods operate on much smaller budgets. GeoThinker uses a fixed number of frames regardless of scene complexity (16, 32 or 64), while CoV~\cite{zhao2026covchainofviewpromptingspatial} adaptively selects frames based on the scene, consuming an average of 18.6 and 20.34 frames on ScanNet++ and InteriorGS, respectively. The CoV frame count consists of three components: 16 fixed frames for initial view selection, 1 Bird's Eye View (BEV) frame, and a small number of additional adaptive frames (2.6 on ScanNet++, 3.34 on InteriorGS). Think3D~\cite{zhang2026think3dthinkingspacespatial} uses 16 uniformly sampled image frames. Besides this it also samples some frames internally from the video.

\begin{table}[t]
\centering
\caption{Existence Question benchmark results categorized by architecture. We report Precision (Prec), Recall (Rec), F1 Score, and Accuracy across ScanNet++ and InteriorGS dataset for different number of frames in GeoThinker.}
\label{tab:existence_geothinker_more_frames}
\small
\setlength{\tabcolsep}{4pt}
\resizebox{\linewidth}{!}{
\begin{tabular}{l cccc c cccc}
\toprule
& \multicolumn{4}{c}{\textbf{ScanNet++}} & & \multicolumn{4}{c}{\textbf{InteriorGS}} \\
\cmidrule(lr){2-5} \cmidrule(lr){7-10}
Method & Prec & Rec & F1 & Acc & & Prec & Rec & F1 & Acc \\
\midrule
GeoThinker (16 frames) & 0.60 & 0.86 & 0.71 & 0.65 & & 0.66 & 0.77 & 0.71 & 0.67 \\
GeoThinker (32 frames) & 0.61 & 0.85 & 0.71 & 0.66 & & 0.61 & 0.79 & 0.69 & 0.63 \\
GeoThinker (64 frames) & 0.62 & 0.84 & 0.71 & 0.67 & & 0.63 & 0.79 & 0.70 & 0.65 \\
\bottomrule
\end{tabular}
}
\end{table}

\begin{table}[t]
\centering
\caption{Spatial Intelligence benchmark results. We report Object Count (Cnt), Object Size (Sz), Absolute Distance (Abs), Relative Distance (Rel-D), and Relative Direction (Rel-Dir) for different number of frames in GeoThinker.}
\label{tab:spatial_geothinker_more_frames}
\small
\setlength{\tabcolsep}{2pt}
\resizebox{\linewidth}{!}{
\begin{tabular}{l ccccc @{\hskip 5pt} ccccc}
\toprule
& \multicolumn{5}{c}{\textbf{ScanNet++}} & \multicolumn{5}{c}{\textbf{InteriorGS}} \\
\cmidrule(lr){2-6} \cmidrule(lr){7-11}
Method & Cnt & Sz & Abs & Rel-D & Rel-Dir & Cnt & Sz & Abs & Rel-D & Rel-Dir \\
\midrule
GeoThinker (16 frames) & 0.19 & 0.57 & 0.33 & 0.47 & 0.51 & 0.24 & 0.48 & 0.16 & 0.28 & 0.36 \\
GeoThinker (32 frames) & 0.27 & 0.59 & 0.33 & 0.52 & 0.53 & 0.27 & 0.50 & 0.22 & 0.27 & 0.40 \\
GeoThinker (64 frames) & 0.29 & 0.58 & 0.35 & 0.51 & 0.54 & 0.30 & 0.50 & 0.20 & 0.29 & 0.40 \\
\bottomrule
\end{tabular}
}
\end{table}

\begin{table}[t]
\centering
\caption{Reasoning benchmark results. We report BLEU-1 (B1), BLEU-4 (B4), CIDEr (C), METEOR (M), and ROUGE-L (R-L) for different number of frames in GeoThinker.}
\label{tab:reasoning_geothinker_more_frames}
\resizebox{\linewidth}{!}{
\begin{tabular}{l ccccc c ccccc}
\toprule
& \multicolumn{5}{c}{\textbf{ScanNet++}} & & \multicolumn{5}{c}{\textbf{InteriorGS}} \\
\cmidrule(lr){2-6} \cmidrule(lr){8-12}
Method & B1 & B4 & C & M & R-L & & B1 & B4 & C & M & R-L \\
\midrule
GeoThinker (16 frames) & 25.98 & 4.77 & 18.19 & 11.52 & 20.90 & & 27.89 & 5.24 & 18.91 & 12.43 & 22.32 \\
GeoThinker (32 frames) & 27.06 & 4.81 & 18.68 & 11.91 & 21.33 & & 27.85 & 5.27 & 20.16 & 12.54 & 22.61 \\
GeoThinker (64 frames) & 27.17 & 4.79 & 19.31 & 12.17 & 21.63 & & 27.65 & 5.09 & 19.84 & 12.71 & 22.68 \\
\bottomrule
\end{tabular}
}
\end{table}

\begin{table}[t]
\centering
\caption{Average number of frames consumed per question across methods on ScanNet++ and InteriorGS datasets. * represents additional sampling from video.}
\label{tab:frames_comparison}
\small
\setlength{\tabcolsep}{4pt}
\begin{tabular}{l c c c c}
\toprule
& \multicolumn{1}{c}{\textbf{ScanNet++}} & & \multicolumn{1}{c}{\textbf{InteriorGS}} \\
\cmidrule(lr){2-2} \cmidrule(lr){4-4}
Method & Frames & & Frames \\
\midrule
\textit{Gemini-Based Models} \\
Video (Basic)          & 165   & & 177   \\
\midrule
\textit{Qwen-Based Models} \\
Video (Basic)          & 330   & & 354   \\
\midrule
\textit{Agentic Methods} \\
CoV                    & 16+1+2.6  & & 16+1+3.34 \\
Think3D                & 16*   & & 16*   \\
\midrule
\textit{Geometry Integrated VLMs} \\
GeoThinker & 16,32,64    & & 16,32,64    \\
GeoThinker (Finetuned) & 16    & & 16    \\
\bottomrule
\end{tabular}
\end{table}

\section{Human Evaluation Interface}
\label{sec:Human Evaluation Interface}
We developed a user rating interface using the viser library to facilitate a seamless evaluation of 3D scene-specific QA pairs. The interface streamlines the assessment process by providing a dynamic 3D environment where users can interactively verify model outputs.

\paragraph{User and Scene Management.}
Upon initializing the interface, users must either create a unique profile or log in to an existing one to track their progress and save ratings (\cref{fig:interface_start}). A dropdown menu allows users to switch between different scenes, which dynamically loads the corresponding Gaussian Splatting data and labels. Once a scene is active, the interface populates a flattened list of questions sampled for that specific environment. Users navigate through these pairs using ``Next Question'' and ``Previous Question'' buttons, which automatically update the 3D visualization, camera focus, and sidebar information.

\paragraph{Visualization and Feedback.}
To assist users in rapid identification, the system overlays object labels and 3D bounding boxes directly onto the scene, employing a color-coded highlighting scheme based on the question type. For existence and count queries, the system highlights target objects in green and contextually relevant objects of the same class in yellow, as illustrated in \cref{fig:interface_existence_and_dir} and \cref{fig:interface_count}. In size and absolute distance tasks, all referred objects appear in green (\cref{fig:interface_distance}). Relative distance questions utilize a cyan box for the reference object, while highlighting the correct choice in green and incorrect choices in red. For relative direction, the interface distinguishes between the origin (blue), target (cyan), and selected (green) objects (\cref{fig:interface_existence_and_dir}). Conversely, for reasoning tasks, the interface provides no automated highlighting; instead, users are encouraged to explore the 3D scene freely to verify the complex logic presented in the sidebar (\cref{fig:interface_reasoning}). After evaluating a pair, users assign a rating from 1 to 5, which is then stored in their local user profile.

\paragraph{Question Distribution.}
The user study involves a curated selection of 100 questions sampled from 10 different scenes. This sample set includes some perturbed samples as well. The specific details regarding task types and their corresponding sampled and perturbed counts are provided in \cref{tab:question_distribution}.

\paragraph{Sample Perturbation.}
To evaluate model robustness, we systematically perturb a subset of the questions across all task categories. For existence queries, we flip the ground truth by changing a ``Yes'' question to ``No'' or vice versa. In counting tasks, we modify the question by adding or removing instances of the target class, or we alter the question's descriptive attributes---for example, changing ``red chair'' to ``black chair''---to invalidate the original count.

For absolute distance and size tasks, we perturb the ground truth by multiplying or dividing the numeric value by a random factor between 5.0 and 10.0. For relative direction and relative distance tasks, we reassign the ground truth to a logically incorrect option. Finally, for reasoning triplets, we utilize a Large Language Model (LLM) to rewrite valid questions into plausible but logically flawed alternatives. While the question text remains unchanged, the answers are rendered incorrect by referencing objects that do not exist in the scene or by modifying the reasoning steps to lead to erroneous answers.

\paragraph{User Study Results.}
The user study included 7 people. As illustrated in the main paper, the users have given the clean samples a high average rating and a consistently better score than the perturbed samples. The difference is particularly noticeable for relative distance and relative direction, questions that humans find easier to answer. The gap between the perturbed and real samples is much smaller for questions like absolute distance and object sizes, which are generally harder questions to answer as they require users to guess.

\begin{table}[h]
\centering
\caption{Distribution of sampled questions for human evaluation across different task types, including the count of perturbed samples.}
\label{tab:question_distribution}
\resizebox{\linewidth}{!}{
\begin{tabular}{lcc}
\toprule
\textbf{Task Type} & \textbf{Total Sampled} & \textbf{Perturbed Samples} \\ \midrule
Existence          & 14                     & 2                         \\
Counting           & 14                     & 3                         \\
Object Size        & 13                     & 3                         \\
Absolute Distance  & 13                     & 3                         \\
Relative Distance  & 13                     & 3                         \\
Relative Direction & 13                     & 3                         \\
Reasoning          & 20                     & 3                         \\ \midrule
\textbf{Total}     & \textbf{100}           & \textbf{20}                \\ \bottomrule
\end{tabular}
}
\end{table}

\section{Qualitative Results}
\label{sec:Qualitative Results}
\begin{figure*}[ht]
    \centering
    \includegraphics[width=\textwidth]{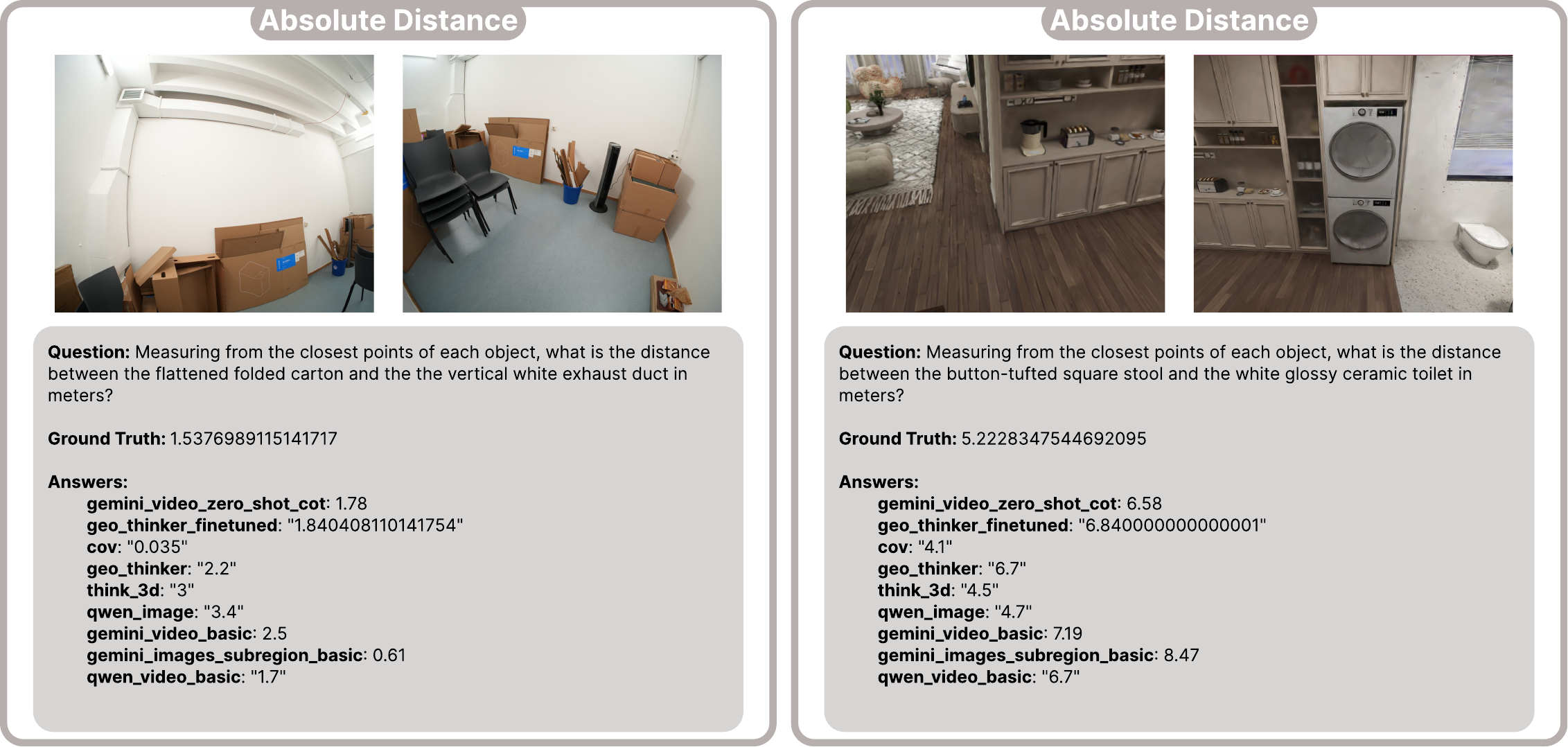}
    \caption{Absolute Distance Question.}
    \label{fig: absolute distance question}
\end{figure*}

\begin{figure*}[ht]
    \centering
    \includegraphics[width=\textwidth]{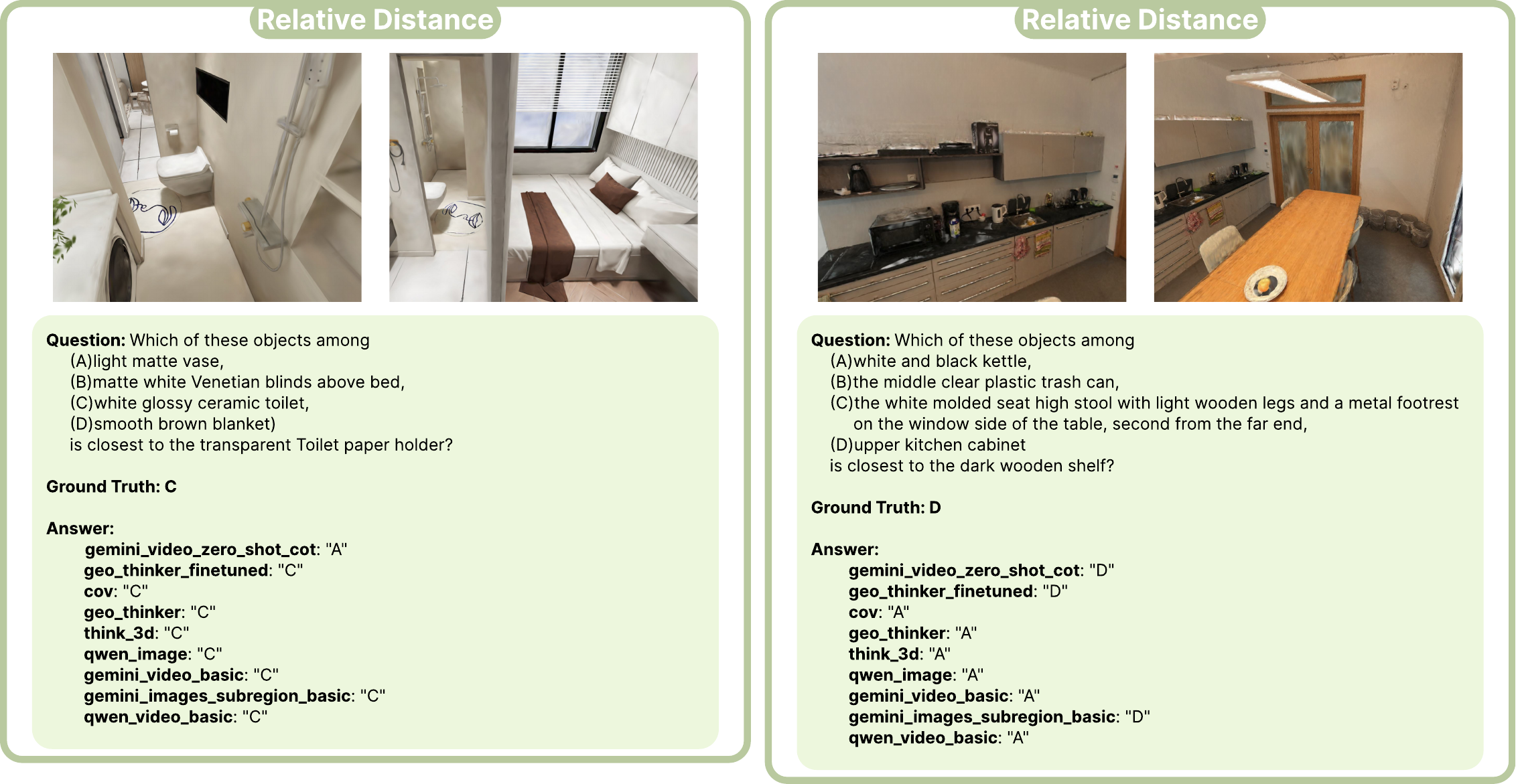}
    \caption{Relative Distance Question.}
    \label{fig: relative distance question}
\end{figure*}

\begin{figure*}[ht]
    \centering
    \includegraphics[width=\textwidth]{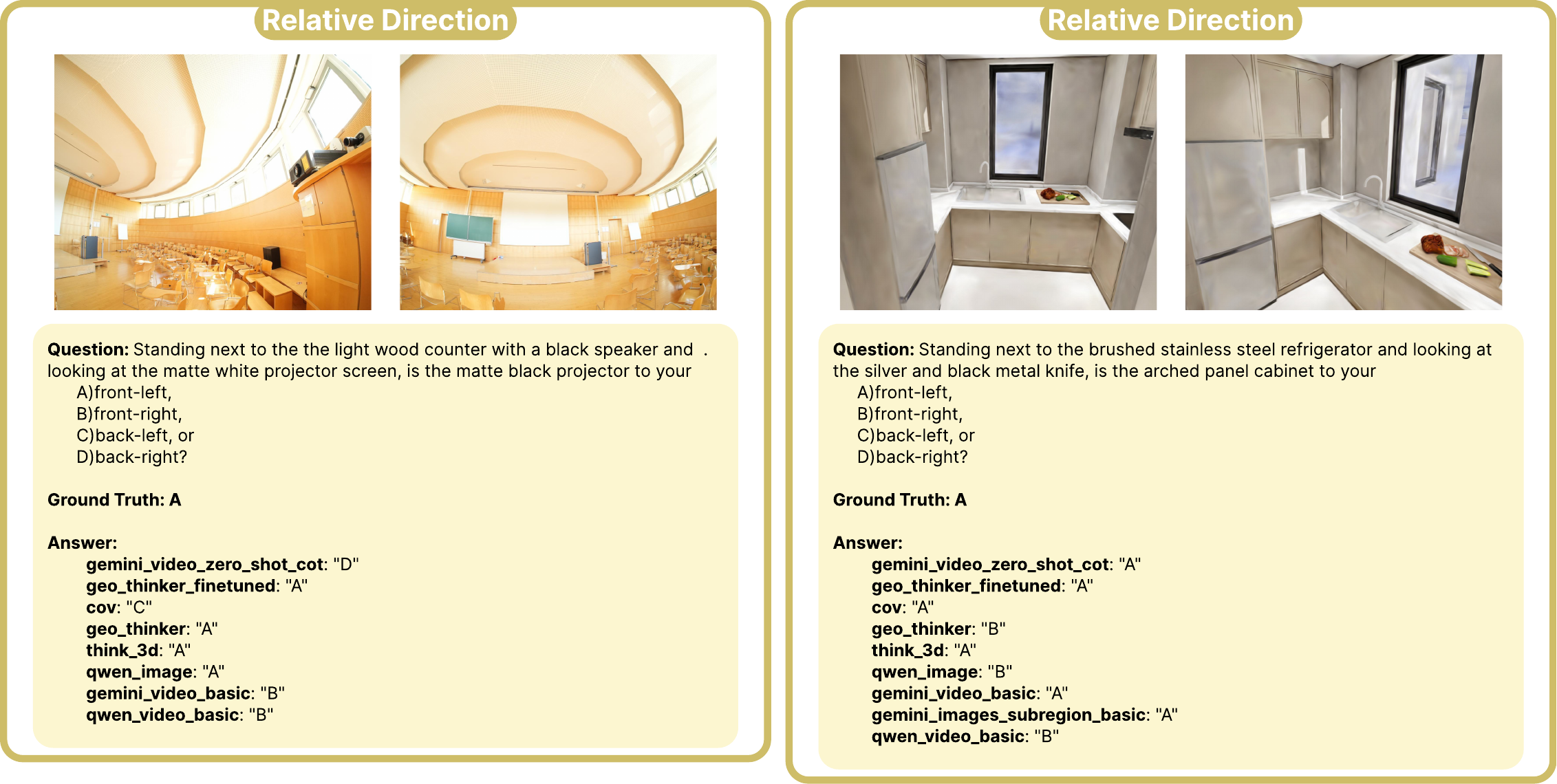}
    \caption{Relative Direction Question.}
    \label{fig: relative direction question}
\end{figure*}

\begin{figure*}[ht]
    \centering
    \includegraphics[width=\textwidth]{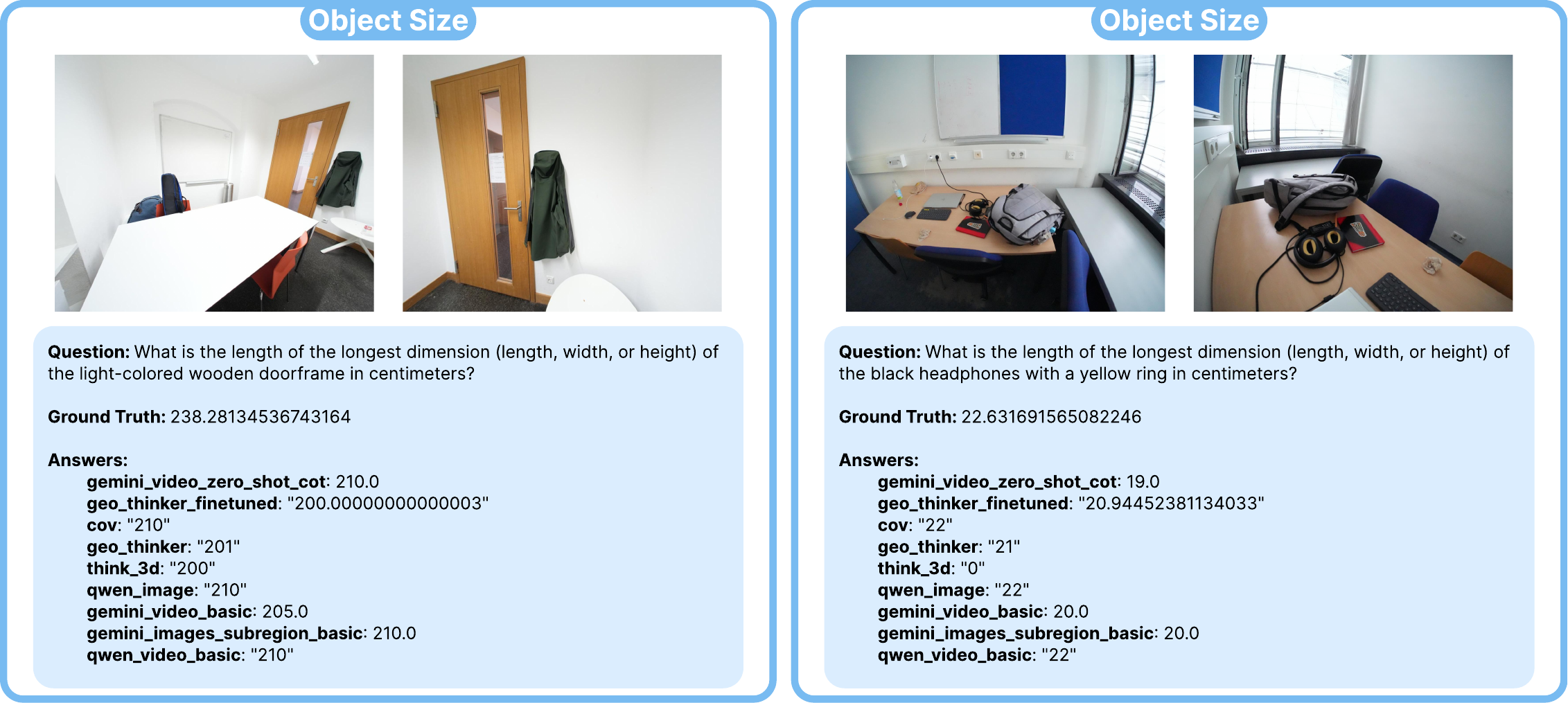}
    \caption{Object Size Question.}
    \label{fig: object size question}
\end{figure*}

\begin{figure*}[ht]
    \centering
    \includegraphics[width=\textwidth]{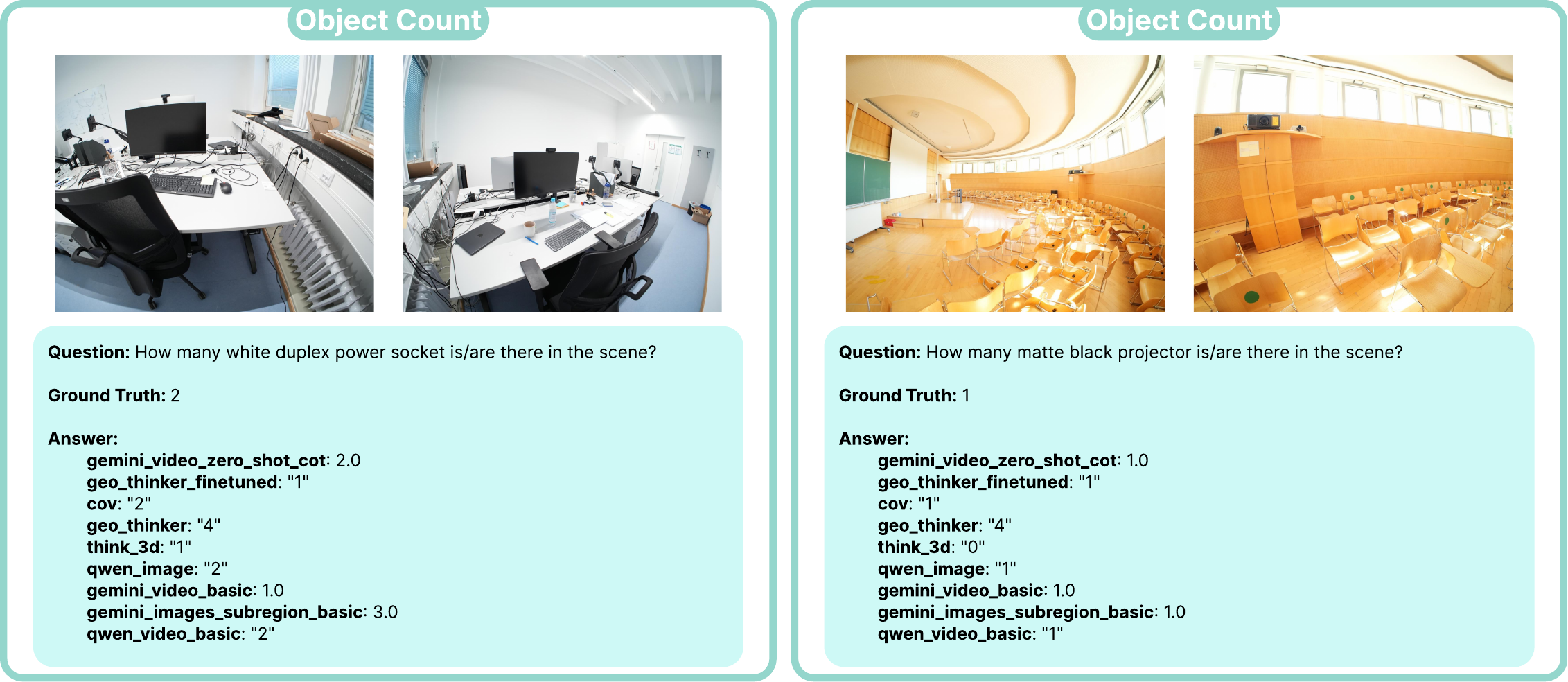}
    \caption{Object Count Question.}
    \label{fig: object count question}
\end{figure*}

\begin{figure*}[ht]
    \centering
    \includegraphics[width=\textwidth]{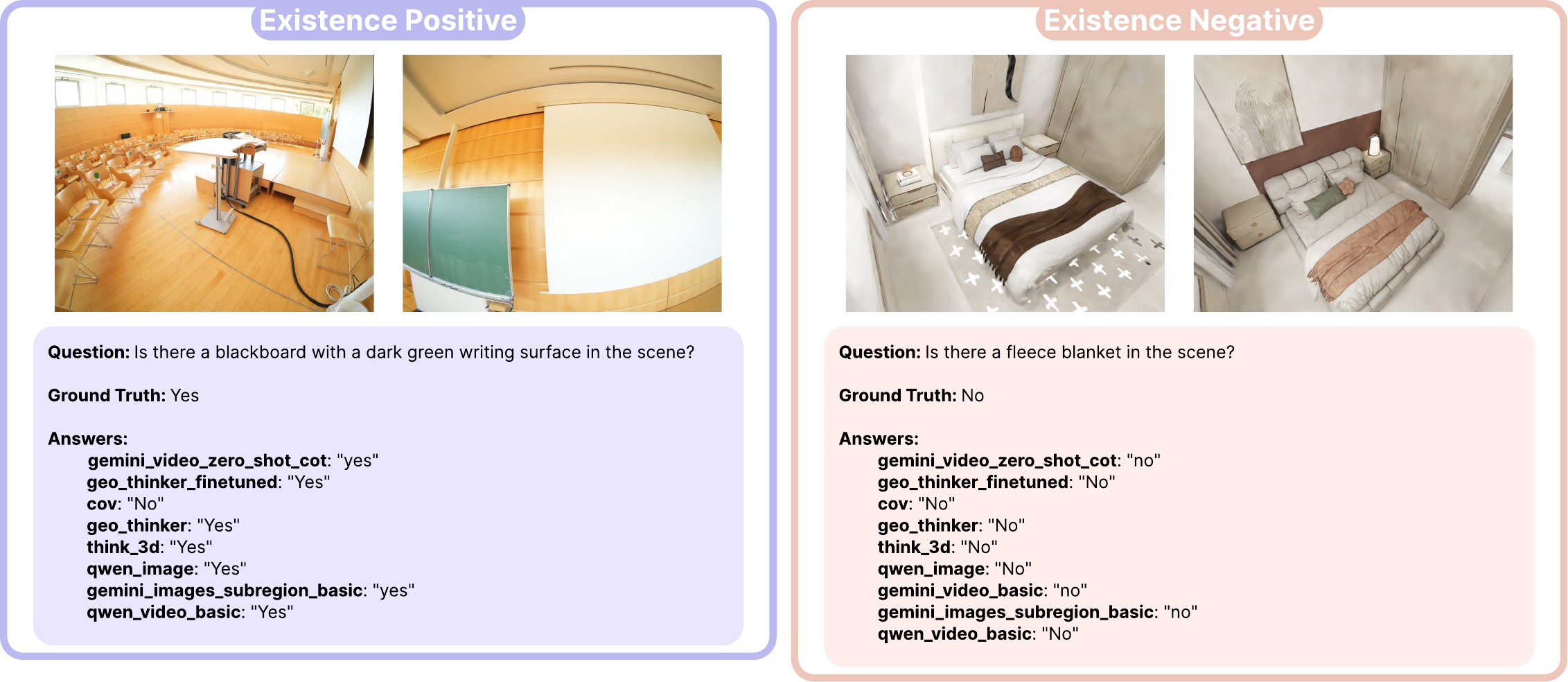}
    \caption{Existence Question.}
    \label{fig: existence question}
\end{figure*}

We present qualitative results for several methods across all task categories. \Cref{fig: existence question} shows the results for existence type questions. \Cref{fig: absolute distance question}, \cref{fig: relative distance question}, \cref{fig: relative direction question}, \cref{fig: object size question} and \cref{fig: object count question} show results for spatial intelligence type questions.

\begin{figure*}[ht]
    \centering
    \includegraphics[width=\textwidth,height=0.95\textheight,keepaspectratio]{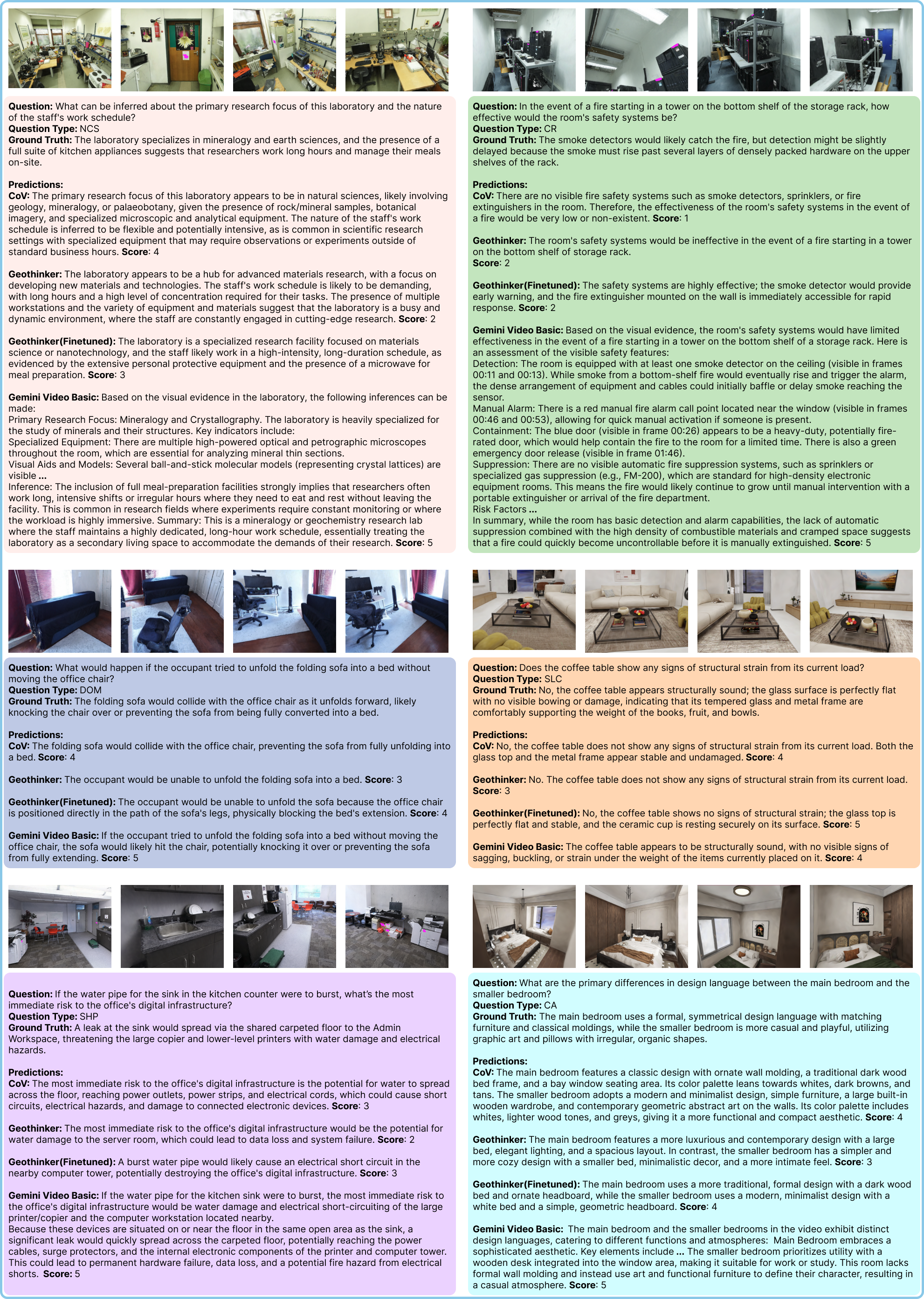}
    \caption{Reasoning Questions.}
    \label{fig: reasoning questions1}
\end{figure*}

\begin{figure*}[ht]
    \centering
    \includegraphics[width=\textwidth,height=0.95\textheight,keepaspectratio]{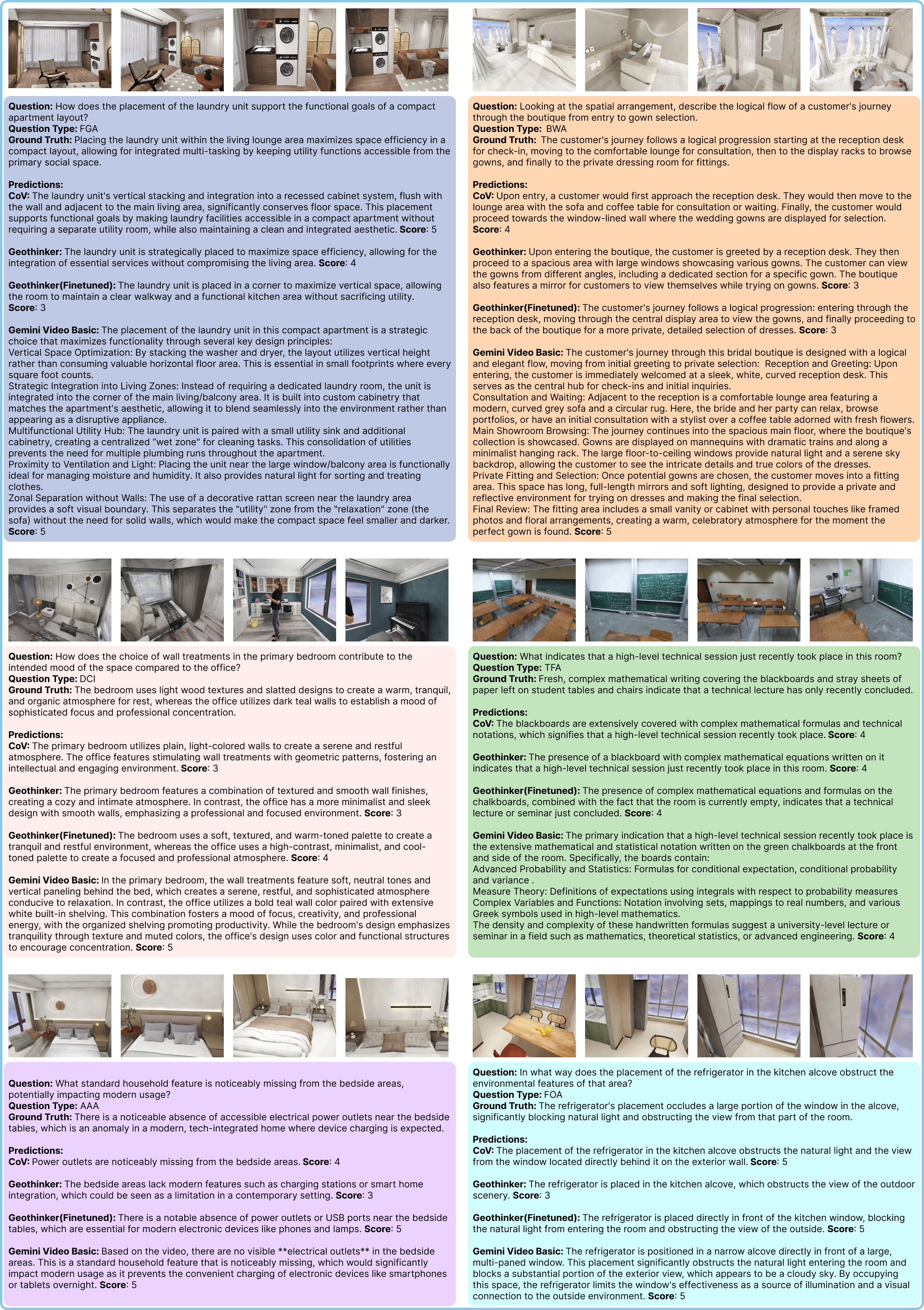}
    \caption{Reasoning Questions.}
    \label{fig: reasoning questions2}
\end{figure*}

\begin{figure*}[t]
    \centering
    \includegraphics[width=\textwidth]{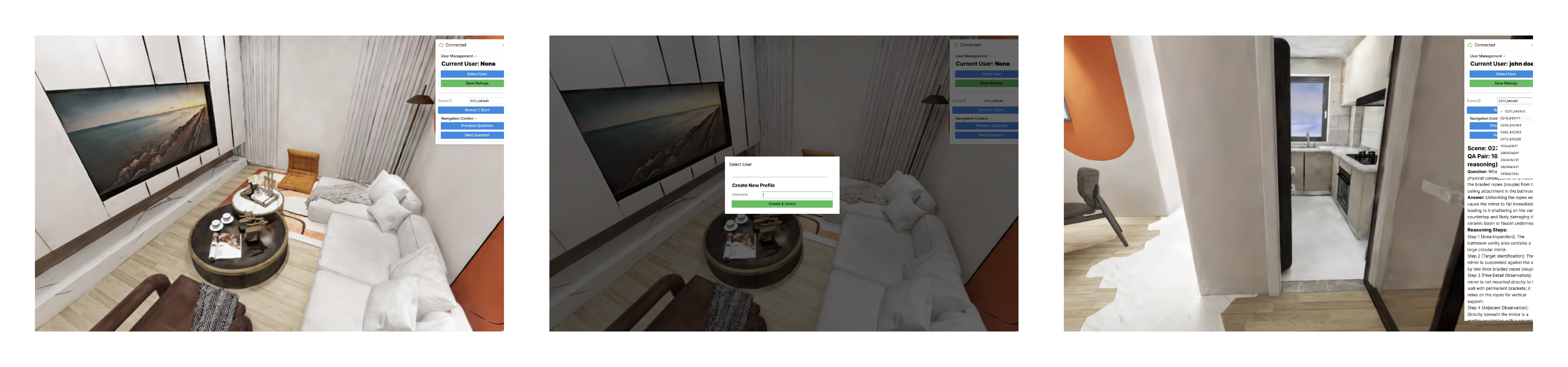}
    \caption{Human Evaluation Interface Overview. The initial setup requires users to create a profile or login to track progress. The interface features a scene selection drop-down and navigation controls to iterate through sampled QA pairs, displaying the question, ground truth answer, and a 1--5 rating scale in the sidebar.}
    \label{fig:interface_start}
\end{figure*}

\begin{figure*}[t]
    \centering
    \includegraphics[width=\textwidth]{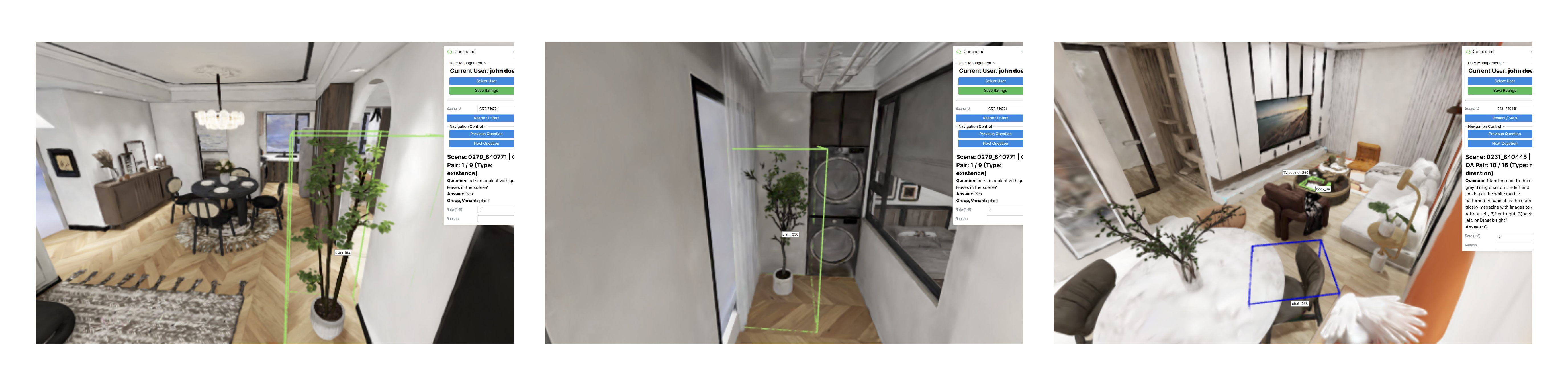}
    \caption{Visualizing Existence and Relative Direction. For existence queries, the system highlights the target object in green; if the object is absent, it highlights other instances of the same class in yellow for context. Relative direction tasks use a color-coded scheme to distinguish between the origin (blue), target (cyan), and selected (green) objects.}
    \label{fig:interface_existence_and_dir}
\end{figure*}

\begin{figure*}[t]
    \centering
    \includegraphics[width=\textwidth]{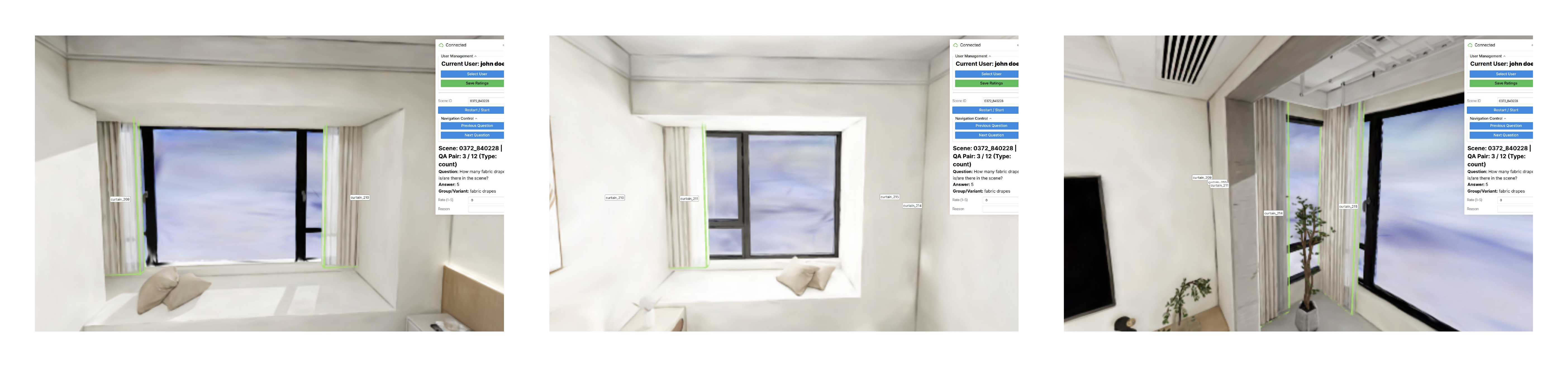}
    \caption{Visualizing Counting Tasks. The interface differentiates object instances based on query attributes. Instances that satisfy the specific requirements mentioned in the question are highlighted with green bounding boxes, while other instances of the same class that do not meet the criteria are highlighted in yellow.}
    \label{fig:interface_count}
\end{figure*}

\begin{figure*}[t]
    \centering
    \includegraphics[width=\textwidth]{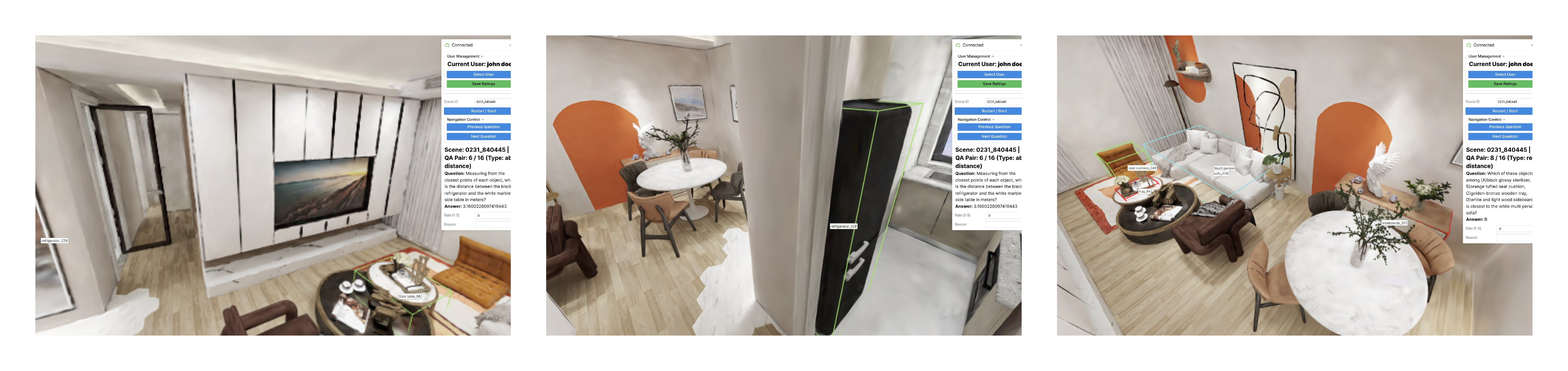}
    \caption{Visualizing Size and Absolute Distance Tasks. All objects referred to in the question are highlighted with green bounding boxes so the annotator can verify the measured dimension or inter-object distance.}
    \label{fig:interface_distance}
\end{figure*}

\begin{figure*}[t]
    \centering
    \includegraphics[width=\textwidth]{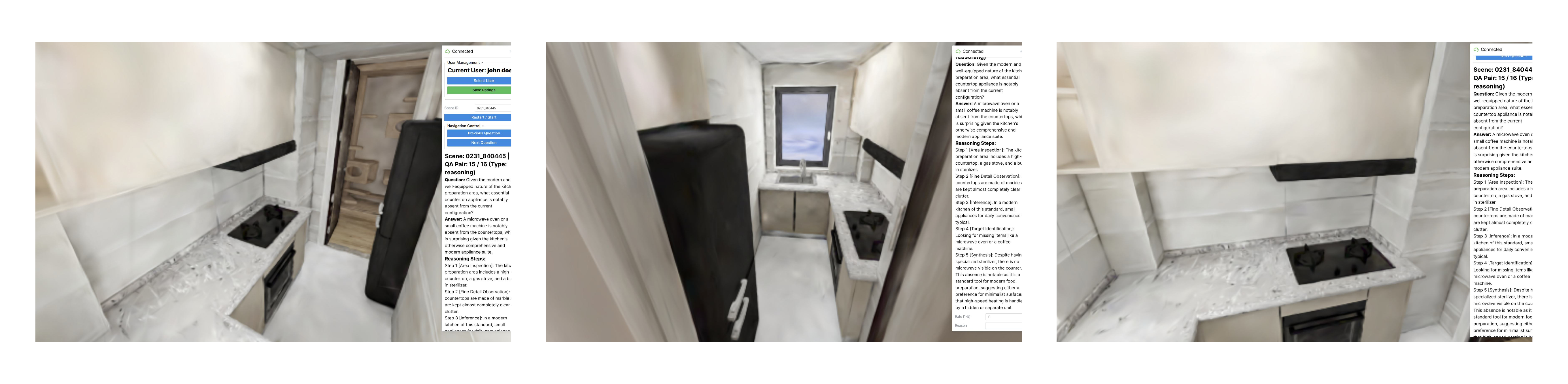}
    \caption{Visualizing Reasoning Tasks. For complex reasoning queries, the interface provides no automated object highlighting. Instead, it displays the multi-step reasoning process in the sidebar and encourages users to explore the 3D scene freely to verify the logic.}
    \label{fig:interface_reasoning}
\end{figure*}

\Cref{fig: reasoning questions1} and \cref{fig: reasoning questions2} present results for the 12 total types of reasoning tasks, displaying one example for each type across the two figures (6 types in each figure). In these reasoning questions, we compare the performance of Gemini Video Basic, CoV, GeoThinker, and GeoThinker-finetuned. For reasoning answers of Gemini involving long text, we use ellipses to maintain conciseness and visual clarity.

\clearpage

\end{document}